\pdfoutput=1

\documentclass{article}

\usepackage[dblblindworkshop, final, nonatbib]{neurips_2025}

\makeatletter
\renewcommand{\@noticestring}{%
  Accepted to the 39th Conference on Neural Information Processing Systems (NeurIPS 2025) Workshop: Learning from Time Series for Health (TS4H).
}
\makeatother

\workshoptitle{Learning from Timeseries for Health (TS4H)}

\usepackage[numbers]{natbib}
\usepackage[utf8]{inputenc}
\usepackage[T1]{fontenc}
\usepackage{amsmath,amsfonts,bm}
\usepackage{graphicx,booktabs,nicefrac,microtype,xcolor,url}
\usepackage{placeins}

\usepackage[hidelinks]{hyperref}
\usepackage[nameinlink,noabbrev]{cleveref}

\usepackage[nameinlink,noabbrev]{cleveref}
\usepackage{caption}

\usepackage{bm}
\usepackage{tabularx}
\usepackage{multicol}
\usepackage{multirow}

\usepackage{subcaption}
\usepackage{threeparttable}
\usepackage{bbm}

\usepackage[none]{hyphenat}

\title{SSM-CGM: Interpretable State-Space Forecasting Model of Continuous Glucose Monitoring for Personalized Diabetes Management}

\author{%
  Shakson ~Isaac\thanks{Both authors contributed equally.} \\
  Department of Biomedical Informatics\\
  Harvard Medical School\\
  \texttt{shakson$\_$isaac@g.harvard.edu} \\
  \And
  Yentl ~Collin$^\ast$\\ 
  Department of Biomedical Informatics\\
  Harvard Medical School\\
  \texttt{yentl.collin@eleves.enpc.fr} \\
  \And
  Chirag J. ~Patel \\
  Department of Biomedical Informatics\\
  Harvard Medical School \\
  \texttt{chirag$\_$patel@hms.harvard.edu}
}

\begin{document}

\maketitle

\begin{abstract}
  Continuous glucose monitoring (CGM) generates dense data streams critical for diabetes management, but most used forecasting models lack interpretability for clinical use. We present SSM-CGM, a Mamba-based neural state-space forecasting model that integrates CGM and wearable activity signals from the AI-READI cohort. SSM-CGM improves short-term accuracy over a Temporal Fusion Transformer baseline, adds interpretability through variable selection and temporal attribution, and enables counterfactual forecasts simulating how planned changes in physiological signals (e.g., heart rate, respiration) affect near-term glucose. Together, these features make SSM-CGM an interpretable, physiologically grounded framework for personalized diabetes management.
\end{abstract}

\section{Introduction}
Diabetes affects over 580 million adults (20–79 years) worldwide \cite{IDF2025Atlas}. Continuous glucose monitoring (CGM) is a central tool for diabetes management, tracking interstitial glucose every 5–15 minutes for up to 2 weeks \cite{DexcomG6}. These data enable metrics such as time-in-range (TIR) which reflect risk of diabetes complications \cite{Zhou2020-ht, Tan2024-cv, Lu2018-zb}. Yet despite these advances, translating CGM data into actionable, personalized guidance remains a central challenge. 

Forecasting models predict near-future glucose trajectories, giving users a “next-hour” view. Existing approaches, including recurrent networks and transformers, demonstrate great predictive accuracy \cite{gluformer2025, glucobench}. However, these studies are limited, showing abilities to predict CGM forecasts but offering little interpretability, limiting usage in personalized, behaviorally-informed care.

We introduce \emph{SSM-CGM}, an interpretable Mamba-based neural state-space model for CGM forecasting and counterfactual reasoning. Using the AI-READI cohort, SSM-CGM: (i) improves forecasting accuracy over the Temporal Fusion Transformer baseline, (ii) provides interpretability via variable selection and temporal attribution, and (iii) generates counterfactual forecasts that simulate how planned changes in physiological indicators influence glucose dynamics. Together, SSM-CGM is a framework that links predictive modeling with the needs of personalized diabetes management.
\section{Methods}
\paragraph{Dataset.}\label{sec:dp} We utilized data from 741 individuals participating in  AI-READI  ~\cite{AI-READI_Consortium2024-gh}, each with 8-10 days of CGM (Dexcom G6) and wearable (Garmin Vivosmart 5) data sampled every 5 minutes, totaling $>$2M measurements (Appendix \ref{app:datasets}).
Wearable features included heart rate, respiration rate, oxygen saturation, steps, stress, calories burned, and sleep stage; demographics are in Appendix \ref{app:datasets}. \textbf{Preprocessing.} Missing values were imputed via Last Observation Carried Forward and series clipped to intervals with overlapping CGM–wearable coverage. \textbf{Feature engineering.} (i) \emph{static covariates}: age, diabetes status, clinical site, and participant ID; (ii) \emph{temporal encodings}: minute-of-day and 24h sine/cosine; (iii) \emph{historical covariates}: recent CGM lags/diffs, rolling stats, time-aligned wearable signals, and meal flags (Sec.~\ref{sec:meal_detection}); (iv) \emph{future covariates}: time-of-day encodings, wearable signals, and meal flags which are proxies for planned daily activities (e.g., sleep, exercise, and eating).

\paragraph{Meal Detection.}\label{sec:meal_detection} As AI-READI lacks meal annotations, we trained a meal-detection model on CGMacros~\cite{Gutierrez-Osuna2025-ug} (Appendix \ref{app:datasets}) using 5-min CGM along 6-h windows. Sparse, time-imprecise labels, were smoothed with a  trapezoidal kernel (30-min plateau, symmetric ramps; $\tilde y_t\!\in\![0,1]$). The model, a dilated CNN + 2-layer Bi-LSTM with LayerNorm and a linear head (Appendix \ref{app:meal-details}), was trained with weighted BCE-with-logits ($\alpha\!=\!5$ to counter class imbalance) on subject-wise, diabetes-stratified train/test splits. At inference on validation series, a sliding 6-h window (stride$=1$) applied per-window thresholding ($\theta\!=\!0.36$) and ratio voting ($\rho\!=\!0.77$). Validation achieved Recall=80\% and Precision=72\%, sufficient for use as a predicted meal flag covariate in AI-READI.

\paragraph{Forecasting Objective.} We predicted future CGM values $\mathbf{y}_{t+1:t+H}$ at 5-minute resolution using the static $\mathbf{s}$, historical $\mathbf{x}^{(\mathrm{hist})}_{t-L+1:t}$, and future $\mathbf{x}^{(\mathrm{fut})}_{t+1:t+H}$ covariates described in Section~\ref{sec:dp}.
\begin{equation}
\hat{\mathbf{y}}_{t+1:t+H}
= f_{\theta}\!\left(
\mathbf{y}_{t-L+1:t},\;
\mathbf{x}^{(\mathrm{hist})}_{t-L+1:t},\;
\mathbf{x}^{(\mathrm{fut})}_{t+1:t+H},\;
\mathbf{s}
\right).
\end{equation}
We used context lengths of $L = 144$ (12\,h), 288 (24\,h), and 576 (48\,h) with a forecast horizon $H=12$ (1\,hour). As participants started and ended device use at different times of day (8-10 days total), we are able to withhold their last two hours of data for validation (penultimate) and testing (last).


\begin{figure}[t]
    \centering 
    \includegraphics[width=0.75\textwidth]{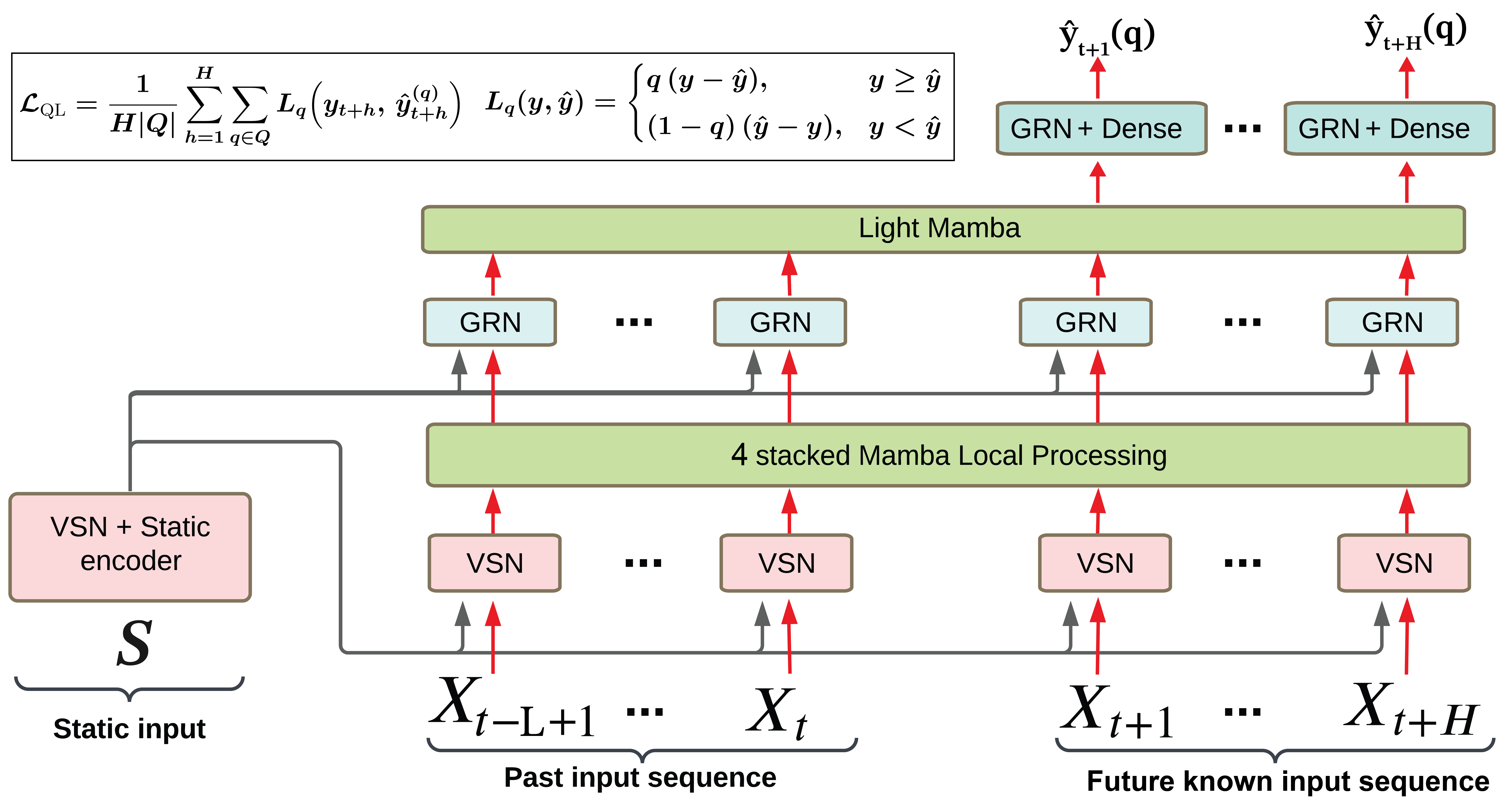}
    \vspace{-0.8em}
    \caption{\textbf{SSM-CGM Architecture.}}
    \label{fig:ModArch}
    \vspace{-1.4em}
\end{figure}
\paragraph{SSM-CGM architecture.}
For our CGM forecasting objective, the \emph{Temporal Fusion Transformer} (TFT)~\cite{Lim2021TemporalFusionTransformers}, often used for CGM forecasting, provides a strong baseline and a great interface: it separates \emph{static}, \emph{historical} and \emph{future} covariates, supports multi-horizon forecasting, and offers interpretable variable importance via \emph{Variable Selection Networks} (VSNs).
In our comparisons, TFT outperformed several strong classical and deep forecasters (LightGBM, N-HiTS, LSTM, etc.) under identical preprocessing (Appendix~\ref{app:quickbenchmark}), consistent with prior CGM studies ~\cite{Zhu2023ISCAS,Zhu2024TBCAS,RodriguezAlmeida2025Sensors,GlucoBench2024}. 

However, CGM dynamics span hours to days, requiring long-context, low-latency models. TFT is widely used; yet sequential LSTMs and $O(L^2)$ attention can be suboptimal for long-context, online forecasting. Recent work shows that Mamba~\cite{Gu2023Mamba}, a selective SSM, provides linear-time scanning, small memory footprint, and a competitive accuracy on time-series benchmarks~\cite{Cai2024MambaTS,Ahamed2024TimeMachine}. Motivated by this, we replace TFT’s LSTM local processor with stacked Mamba blocks and its global self-attention with a light Mamba layer, while retaining the original embeddings, GRNs (\Cref{app:grn}), VSNs (\Cref{app:vsn}), and the multi-horizon head. The resulting architecture (Figure~\ref{fig:ModArch}) preserves TFT’s variable-wise fusion and leverages Mamba’s state-space temporal modeling. The model uses a multi-horizon prediction head trained with a quantile loss (Figure~\ref{fig:ModArch}) $\mathcal{L}_{\text{QL}}$ at $q\in Q=\{0.1,\,0.5,\,0.9\}$ to produce calibrated prediction intervals. 

\paragraph{Interpretability.}\label{sec:interpretability method}  We provide two views of model interpretability in forecasting: (i) \emph{variable importance} from VSNs (“what”), and (ii) \emph{temporal importance} from SSM dynamics (“when”; \emph{hidden attention}). \textbf{Variable importance.} The VSN outputs simplex weights \(\alpha_{i,t}\) over time series covariates and \(\alpha_k^{s}\) over static covariates \(c\) (see Appendix \ref{app:vsn}). Aggregating these weights across participants and contexts reveals which signals drive the forecasts. \textbf{Temporal importance.} Inspired by the \emph{Hidden Attention of Mamba}~\cite{Ali2024HiddenAttentionMamba}, we adapt hidden attention to time-series forecasting by unrolling the linear state update (Eq.~\ref{eq:ssm-evolve}) of SSM equations (see Appendix \ref{app:mamba}). 
This yields a causal convolution with kernel $K_{\ell j}$, inducing attention-like weights over past lags: at time $\ell$, with the input $x$, the output (readout) $z$ and the SSM dynamics $B,C,\Delta,\Lambda$ (see Appendix \ref{app:mamba})
\vspace{-1.6em}

\begin{equation}\label{eq:hidden-ssm}
z_\ell = \sum_{j\le \ell} K_{\ell j}x_j,\quad
K_{\ell j} = C_\ell \exp\!\big(\textstyle\sum_{t=j}^{\ell-1}\Delta t_t\,\Lambda_t\big) B_j,\quad
\alpha_{\ell j} = \operatorname*{softmax}_{j\le \ell}\!\left(\frac{\log(\psi(K_{\ell j})+\varepsilon)}{\tau}\right),
\end{equation}

\vspace{-1.2em}

where $\alpha_{\ell j}$ are attention-like lag weights ($\varepsilon,\tau>0$; $\psi$ is the channel-wise $\ell_{1}$ norm). We further employ \emph{Mamba-2}~\cite{Dao2024TransformersAreSSMs}, which factorizes channels into multi-head groups with stable parameterization, improving throughput and long-range capture. To make head-wise attributions comparable, taking inspiration from TFT’s interpretable attention and the multi-expand SSM (MES) design in \emph{Transformers Are SSMs}~\cite{Dao2024TransformersAreSSMs}, we created an MES-style Mamba2: each head has its own dynamics $(B^{(h)},\Lambda^{(h)},\Delta t^{(h)})$, but we enforce a \emph{shared} value projection $V$ and readout $C_{\mathrm{sh}}$ across heads (see \Cref{app:mamba2}); in practice,  we first apply $v_j = V x_j$ \emph{(shared across all heads)}, then specialize the dynamics per head while keeping the readout fixed to $C_{\mathrm{sh}}$  ($K_{\ell j} \rightarrow K^{(h)}_{\ell j}$). With this change, hidden-attention maps $\alpha^{(h)}_{\ell j}$ are on a common scale, analogous to interpretable multi-head attention in TFT, and empirically yields smoother, more stable temporal attributions.

\paragraph{Counterfactual Forecasting.}\label{sec:counterfactual} We model counterfactual CGM forecasts as interventions on planned future covariates (e.g., “what-if heart rate increases?”). Let $\mathcal{H}_t = (\mathrm{CGM}_{t-L+1:t}, X_{t-L+1:t}, S)$ denote history up to $t$, where $X$ time-varying (e.g., activity, meals) and $S$ static covariates. For a planned future covariate sequence $\mathbf{a}_{t+1:t+h}$, the counterfactual distribution is
\begin{equation}
\label{eq:counterfact}
p\!\left(\mathrm{CGM}_{t+1:t+h}(\mathbf{a}) \mid \mathcal{H}_t\right)
= \mathbb{P}\!\left(\mathrm{CGM}_{t+1:t+h} \mid \mathcal{H}_t, \mathrm{do}(X_{t+1:t+h}=\mathbf{a})\right),
\end{equation}
where $\mathrm{do}(\cdot)$ fixes covariates $\mathbf{a}_{t+1:t+h}$ irrespective of natural dynamics. Under standard assumptions of consistency, sequential ignorability, and positivity, this distribution is identified by the sequential g-formula~\cite{robbins1986gcomp,wang2020gformula} (\Cref{app:plausibility}). In practice, \emph{SSM-CGM} implements this directly by conditioning on the planned sequence in a single pass, rather than computing the g-formula explicitly.

\section{Results}

\begin{table}[b]
\centering
\scriptsize
\vspace{-2.5em}
\caption{Benchmarking performance (1\,hour horizon). Mean (95\% CI).}
\label{tab:bench-mini}
\begin{tabular}{lcccccc}
\toprule
\multirow{2}{*}{Context Length} & \multicolumn{3}{c}{\textbf{SSM-CGM}} & \multicolumn{3}{c}{\textbf{TFT}} \\
\cmidrule(lr){2-4}\cmidrule(lr){5-7}
 & \textbf{Quantile Loss} $\downarrow$ & \textbf{MAE} $\downarrow$ & \textbf{RMSE} $\downarrow$ & \textbf{Quantile Loss} $\downarrow$ & \textbf{MAE} $\downarrow$ & \textbf{RMSE} $\downarrow$ \\
\midrule
12\,h & 3.87 (3.81, 3.93) & 5.97 (5.68, 6.26) & 7.17 (6.83,7.52) & 3.95 (3.89, 4.02) & 6.07  (5.76,  6.37) & 7.32  (6.96, 7.69) \\
24\,h & 3.69 (3.63, 3.75) & 5.66 (5.39, 5.93) & 6.86 (6.53, 7.19) & 3.92 (3.86, 3.99)  & 6.06 (5.77, 6.36)  & 7.28 (6.93, 7.63) \\
48\,h & \textbf{3.52} (3.47, 3.58) & \textbf{5.40} (5.15, 5.65) & \textbf{6.56} (6.26, 6.87) & 3.70 (3.64, 3.76) & 5.77 (5.50, 6.03) & 6.92 (6.61, 7.24) \\
\bottomrule
\end{tabular}
\end{table}

\paragraph{Benchmarking.} We evaluated whether \textit{SSM-CGM} improved forecasting performance compared to its baseline architecture TFT \cite{Lim2021TemporalFusionTransformers}, specifically testing the replacement of TFT’s LSTM/self-attention with our Mamba-2 (MES-style) blocks.  Under identical preprocessing, splits, hyperparameter search, and training on 4 A100 GPUs (see \Cref{app:benchmark}), along with matched parameter counts (SSM-CGM \textit{1,762,714} params vs.\ TFT \textit{1,793,376}), SSM-CGM \textit{outperformed} TFT on the AI-READI dataset (Table \ref{tab:bench-mini}). We report the training objective, MAE, and RMSE to assess sensitivity to large errors. With 12h, 24h, and 48h contexts, SSM-CGM consistently reduces MAE compared to TFT by approximately 2\%, 7\%, and 7\%, respectively. Within SSM-CGM, we observe 5\% performance gain with each doubling of context length, highlighting the value of long-range temporal information in CGM forecasting and the ability of our model to capture it.

\vspace{-0.5em}
\paragraph{Interpretability.}\label{sec:interpret} The VSNs report global and step-wise importances: averages over time/participants (Figure~2A) indicate that past/future wearable covariates strongly help predictions, especially meal flags, stress levels, heart/respiration rates reflecting circadian and activity patterns \cite{CircadianGluc2009,CGMdiet2025NatMed}, while step-wise maps (\Cref{app:interp-vsn}) pinpoint which signals matter at each instant. Static covariates also contributed (Appendix~\ref{app:interp-static}), and participant-level embeddings showed that SSM-CGM captured markers of hyperglycemia and cardiovascular risk (Appendix~\ref{app:embeddings}). Averaging head-wise hidden attention maps 
(Section \ref{sec:interpretability method}) yielded a stable temporal attribution profile, quantifying how past lags influence decoder steps at the cohort and individual level (Figure 2B). As expected of smooth CGM dynamics, recent history dominated, but individuals showed distinctive peaks at specific lags indicating moments the model must retain to improve forecasts. Self-hidden attention maps (Figure 2C, \Cref{app:interp-hidden}) further demonstrated how salient past events propagate through the SSM state and shape downstream predictions. At the cohort level, averaging hidden-attention maps across layers and heads suggests a complementary division of temporal labor in differents mamba layers (Appendix~\ref{app:interp-hidden}). Together, these analyses revealed which covariates matter and when past information is most influential, capturing both cohort-level patterns and individual-specific signals relevant for glucose management.
\vspace{-1.0em}

\begin{figure}[t]
    \centering 
    \includegraphics[width=0.92\textwidth]{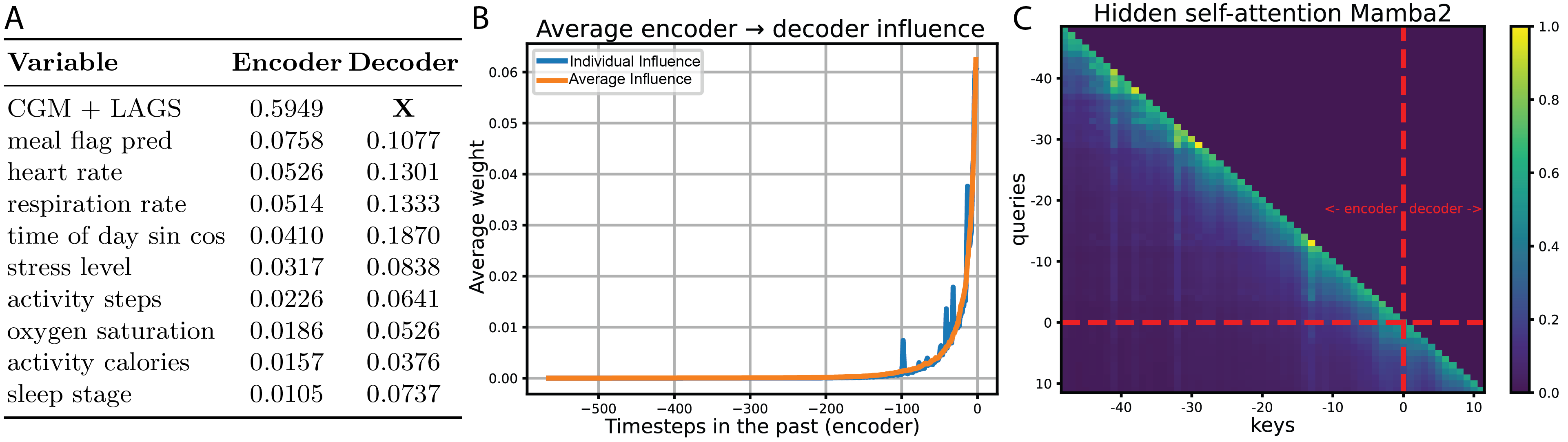}
    \vspace{-0.8em}
    \caption{\textbf{Interpretability of SSM-CGM.}
    (A) VSN average features importance (B) Past influence on predictions (average, individual) (C) Head average self-attention map of Light Mamba (last 60-step window, one individual)
    }
    \label{fig:interpretable}
    \vspace{-1.5em}
\end{figure}

\begin{figure}[ht]
    \centering 
    \includegraphics[width=0.92\textwidth]
    {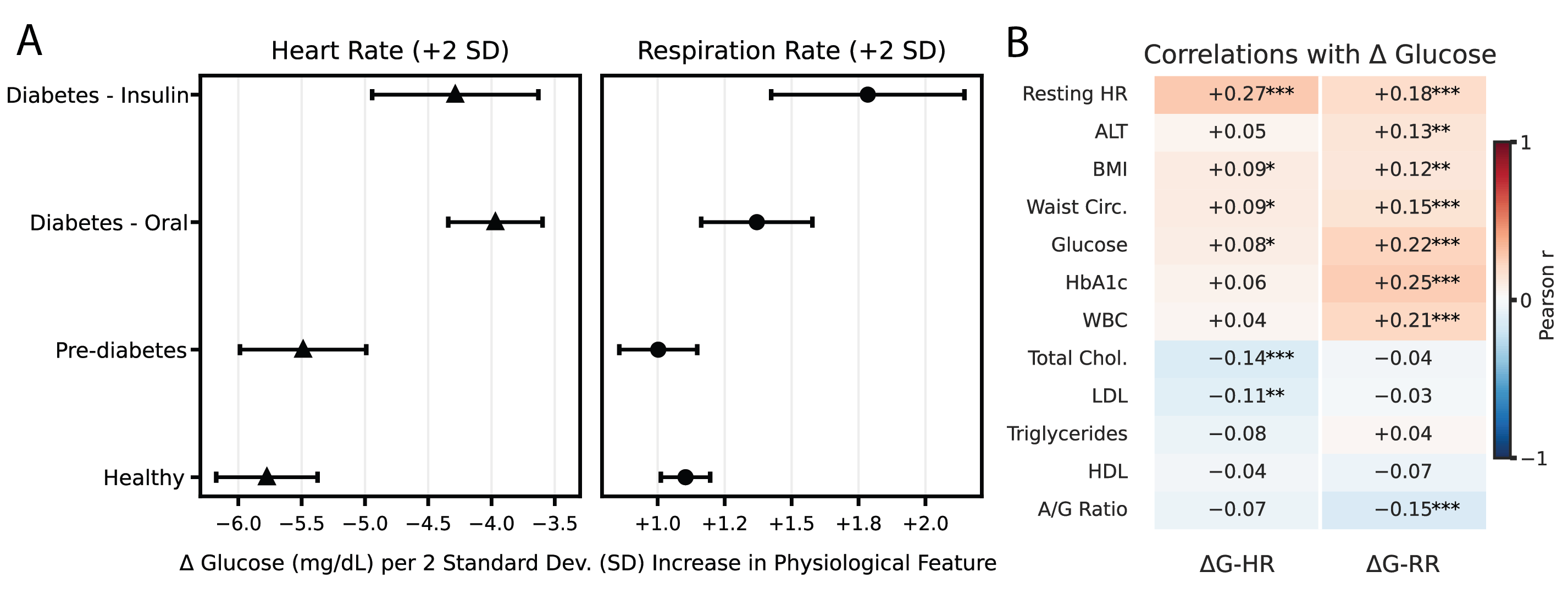}
    \vspace{-1.1em}
    \caption{\textbf{Counterfactual effects and associations.}
    (A) Estimated change in glucose (mg/dL) from a +2\,SD perturbation to heart rate (HR) or respiration rate (RR) over a 1-h horizon, stratified by diabetes status (mean $\pm$ 95\% CI).
    (B) Pearson correlations between the counterfactual effects ($\Delta$G-HR, $\Delta$G-RR) and clinical laboratory measures. Stars indicate FDR-adjusted significance (p): * $<\!0.05$, ** $<\!0.01$, *** $<\!0.001$.}
    \label{fig:counterfact}
\end{figure}
\vspace{-1.40em}
\paragraph{Counterfactuals.}
We next evaluated whether short-term interventions in planned activity could alter glucose trajectories. Using SSM-CGM, we computed counterfactuals (Section~\ref{sec:counterfactual}), for the two top-ranked wearable features: heart rate (HR) and respiration rate (RR) (Figure \ref{fig:counterfact}). Each counterfactual estimated change-in glucose ($\Delta$ glucose) when the feature is perturbed to +2 within-person SD above the individual's baseline mean value across 1-hour horizons restricted to plausible times of day (Appendix~\ref{app:plausibility}). On average, the maximal short-term effect on glucose was \(-5.0\) mg/dL for HR and \(+1.2\) mg/dL for RR. Because counterfactuals are computed at the individual level, we examined their variation across diabetes status (Figure 3A). We found that $\Delta$G-HR and $\Delta$G-RR were uncorrelated (r = 0.01) but mapped to distinct domains of health: HR effects correlated with cardiovascular and glycemic markers, while RR effects aligned strongly with adiposity, inflammation, and hepatic dysfunction (Figure 3B). Importantly, the simulated effects reflected plausible contexts. Daytime HR increases, consistent with physical activity, were associated with improved glucose tolerance. In contrast, nighttime RR increases, mimicking impaired sleep respiration, were linked to adverse metabolic markers.

\section{Conclusion}

SSM-CGM advances CGM forecasting with interpretable, counterfactual, and personalized insights for diabetes management. Despite limitations (Appendix~\ref{app:limitations}), this work is an initial step toward deep models that not only forecast CGM trajectories but also guide individualized behavioral plans to improve outcomes such as time-in-range.


\newpage

\appendix

\crefalias{section}{appendix}
\crefalias{subsection}{appendix}
\crefalias{subsubsection}{appendix}
\Crefname{appendix}{Appendix}{Appendices}
\crefname{appendix}{appendix}{appendices}

\section{Technical Appendices and Supplementary Material}

\subsection{Datasets}\label{app:datasets}

The Artificial Intelligence Ready and Exploratory Atlas for Diabetes Insights (AI-READI) is a multimodal, curated dataset of approximately 1,000 individuals spanning glycemic states from normoglycemia to prediabetes, type 2 diabetes managed with oral medications, and type 2 diabetes managed with insulin \cite{AI-READI_Consortium2024-gh}. Inclusion criteria required participants to be over 40 years old and living with or without type 2 diabetes, while exclusion criteria specified that individuals with gestational diabetes or type 1 diabetes were not eligible. Summarized statistics for AI-READI, including demographics, clinical categories, data characteristics, and CGM-derived metrics, are provided in Table~\ref{tab:cohort_summary}. Sex, race, and ethnicity were not available, as these sensitive variables are excluded from the public AI-READI release to reduce re-identification risk and potential stigmatization. For our analyses, we used 741 individuals with CGM, wearable activity, and clinical laboratory features from AI-READI.

The CGMacros dataset consists of 45 adults across normoglycemia, prediabetes, and type 2 diabetes, combining CGM with detailed dietary information \cite{Gutierrez-Osuna2025-ug}. Participants were monitored for approximately 10 days, during which they wore a Dexcom G6 CGM and logged descriptions and photographs of their meals, enabling the estimation of caloric intake and macronutrient composition. Summarized statistics for CGMacros, including demographics, clinical characteristics, data features, and CGM-derived metrics, are also presented in Table~\ref{tab:cohort_summary}.

The similarities in both datasets such as demographics, clinical features, and CGM-derived metrics gave us confidence to train a meal detection model in CGMacros that can be deployed on the AI-READI dataset.

\begin{table}[ht]
\centering
\caption{Cohort characteristics for AI-READI and CGMacros. Values are mean $\pm$ SD unless otherwise indicated.}
\label{tab:cohort_summary}
\begin{threeparttable}
\begin{tabular}{lcc}
\toprule
\textbf{Characteristic} & \textbf{AI-READI} & \textbf{CGMacros} \\
\midrule
\multicolumn{3}{l}{\textbf{\textit{Demographics}}} \\
\quad Sample size ($N$)         & 741                 & 45 \\
\quad Age (years)                & 60 $\pm$ 11         & 48 $\pm$ 13 \\
\quad BMI (kg/m$^2$)             & 30.9 $\pm$ 8.4      & 31.1 $\pm$ 6.7 \\
\midrule
\multicolumn{3}{l}{\textbf{\textit{Clinical}}} \\
\quad Normoglycemia              & 275 (37\%)          & 15 (33\%) \\
\quad Prediabetes                 & 163 (22\%)          & 16 (36\%) \\
\quad T2D, non-insulin           & 213 (29\%)          & 14 (31\%) \\
\quad T2D, insulin-treated       & 90 (12\%)           & Not assessed \\
\quad HbA1c (\%)                 & 6.0 $\pm$ 1.3       & 6.1 $\pm$ 0.9 \\
\midrule
\multicolumn{3}{l}{\textbf{\textit{Data characteristics}}} \\
\quad CGM device                 & Dexcom G6           & Dexcom G6 \\
\quad Wearable device            & Garmin Vivosmart 5  & Not collected \\
\quad Sampling interval          & 5 min               & 5 min \\
\quad Recording duration (days)  & 9.6 $\pm$ 0.8       & 9.7 $\pm$ 0.9 \\
\quad Total measurements         & 2,045,164            & 123,318 \\
\midrule
\multicolumn{3}{l}{\textbf{\textit{CGM-derived metrics}}} \\
\quad TIR (70--180 mg/dL, \%)    & 85.8 $\pm$ 22.1     & 85.6 $\pm$ 20.0 \\
\quad Mean glucose (mg/dL)       & 138.6 $\pm$ 40.3    & 140.4 $\pm$ 28.3 \\
\quad CV (\%)                    & 19.6 $\pm$ 5.6      & 20.1 $\pm$ 5.4 \\
\bottomrule
\end{tabular}
\begin{tablenotes}
\small
\item T2D = type 2 diabetes; CGM = continuous glucose monitoring; TIR = time in range (70–180 mg/dL); CV = coefficient of variation.
\end{tablenotes}
\end{threeparttable}
\end{table}

\newpage
\subsection{Meal Detection (details complementing Sec.~\ref{sec:meal_detection})}\label{app:meal-details}

To supply meal covariates for AI-READI, which lacks meal annotations, we trained a meal-detection model on the PhysioNet CGMacros dataset~\cite{Gutierrez-Osuna2025-ug} using 5-min CGM sampled into 6-h windows.

\paragraph{Problem setup.}
We frame meal detection as a binary event detection task from CGM-only time series sampled every 5 minutes. The signal is processed in sliding 6-hour windows ($L{=}72$ samples). For each timestep $t$ in a window, the model outputs a probability $\hat p_t\in[0,1]$ that a meal occurs at that moment.
Because the ground-truth meal times are time-uncertain and the data are class-imbalanced (few meal minutes vs. many non-meal minutes), we convert point annotations into smoothed targets (See Trapezoidal label smoothing section). At inference, we convert the probabilistic sequence into a binary meal-flag sequence $\hat y_t\in{0,1}$ by thresholding within each window and reconciling overlaps with a local consensus rule (ratio voting), yielding a single timeline of detected meal events.

\begin{figure}[h]
\centering
\includegraphics[width=\linewidth]{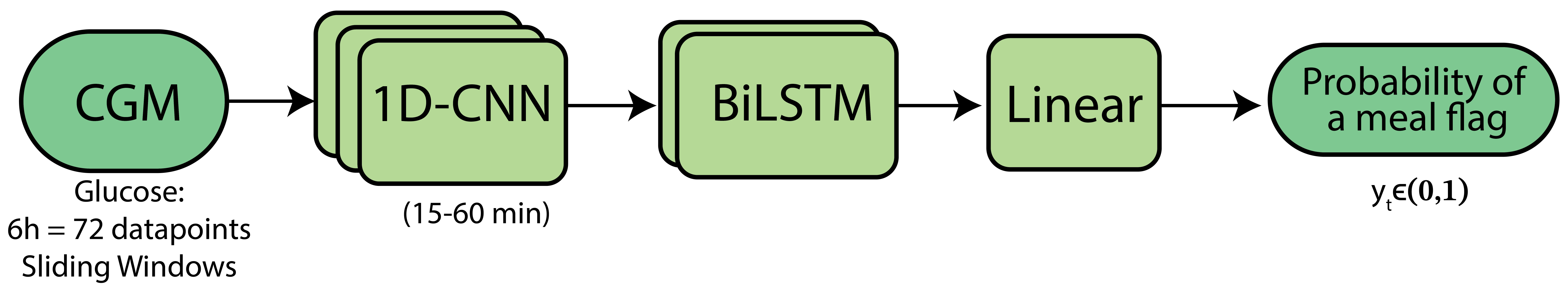}
\caption{Meal detection model architecture}
\label{fig:meal-arch}
\end{figure}

\paragraph{Model recap.}
We use a dilated 1D CNN (kernel sizes $15/3/3$; dilations $1/2/4$) to capture short/medium motifs, followed by a 2-layer BiLSTM ($h{=}64$ per direction) for temporal context, LayerNorm, dropout ($0.4$), and a linear head producing per-step logits (Figure \ref{fig:meal-arch}). The loss is a weighted BCE-with-logits: 

\[
\mathcal{L}
= -\frac{1}{L}\sum_{t=1}^{L}\!\left[
\alpha\,\tilde y_t \log \sigma(\hat y_t) + (1{-}\tilde y_t)\log\!\bigl(1-\sigma(\hat y_t)\bigr)
\right],
\]

where $\alpha{=}5$ upweights positives to counter class imbalance, $\sigma$ is the sigmoid, and $\tilde{y}_t$ the smoothed target ($\tilde{y}_t \in [0, 1]$). 

\emph{Training Setup:} Sampling 5-minute CGM; window length $L{=}72$; stride $s{=}1$; CNN channels $\texttt{cout}{=}16$; 2 BiLSTM hidden $h{=}64$ per direction; dropout $0.4$; LayerNorm; Adam (lr $10^{-4}$, weight decay $10^{-4}$); OneCycleLR; batch size $16$; $10$ epochs; positive class weight $\alpha{=}5$; subject-wise splits (no leakage) stratified by diabetes status (healthy, prediabetes, diabetes).

\paragraph{Trapezoidal label smoothing.}
Each annotated meal time $\tau_k$ is expanded into a trapezoid
$w(t;\tau_k,P,R)$ with a central plateau (target $=1$) and linear ramps (decay to $0$):
\[
w(t;\tau_k,P,R)=
\begin{cases}
0, & |t-\tau_k|\ge P{+}R,\\
1, & |t-\tau_k|\le P,\\
1-\dfrac{|t-\tau_k|-P}{R}, & \text{otherwise}.
\end{cases}
\]

Intuition: the plateau (here $\approx 20$\,min total) tolerates small timing errors, while linear ramps encode decreasing responsibility farther from the reported meal. Accordingly, in our implementation we set $P{=}2$ (plateau length $\approx 10$ min) and $R{=}2$ (ramp length $\approx 10$ min). Within each 6\,h window ($L{=}72$), we aggregate all trapezoids that overlap the window
by summation and clip to keep targets bounded for BCE:
\[
S_t \;=\; \sum_{k} w(t;\tau_k,P,R), \qquad
\tilde y_t \;=\; \min\!\bigl(1,\, S_t\bigr).
\]
In practice, meal overlaps are rare so the clipping is almost never
activated and has negligible impact on supervision. (Alternative bounded unions such as
$\max_k w(\cdot)$ behave similarly in our setting.)

\begin{figure}[h]
\centering
\includegraphics[width=0.8\linewidth]{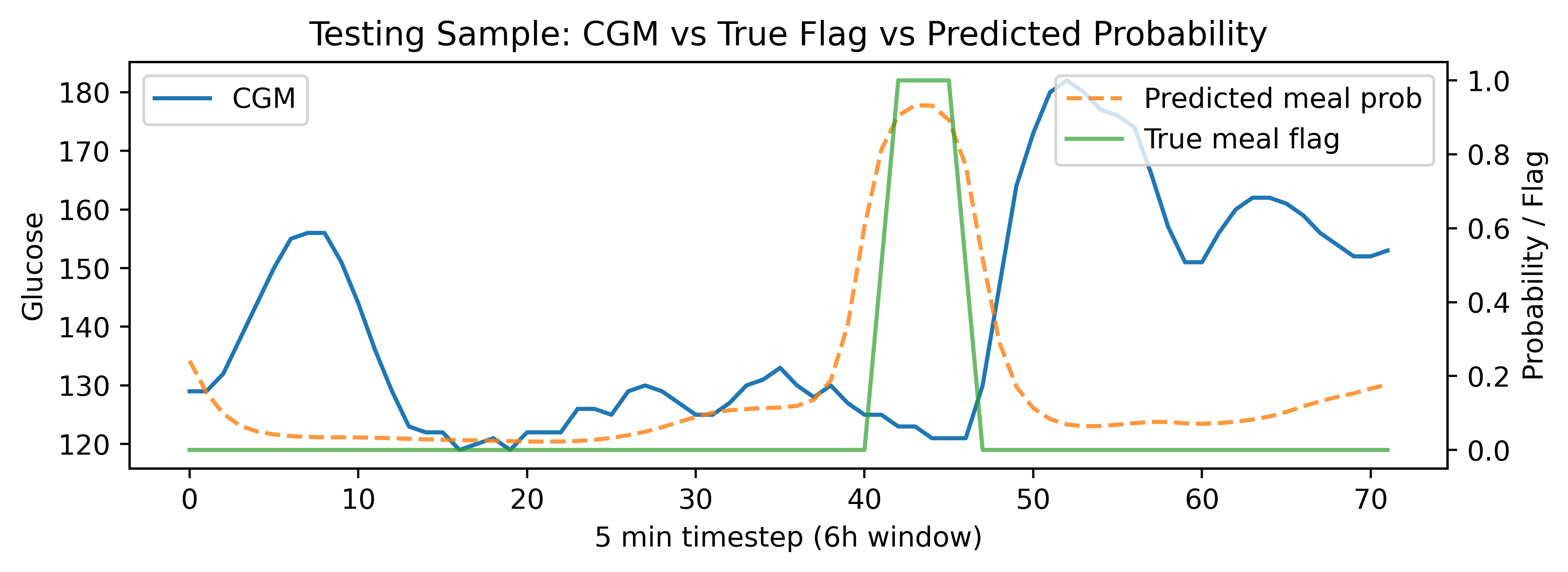}
\caption{Predicted meal flag probability on test set window}
\label{fig:mealpredprob}
\end{figure}

\paragraph{Inference \& reconstruction (thresholding + ratio voting).}
At test time, we slide windows with stride $1$ and (i) apply a \emph{per-window threshold} $\theta$ to convert probabilities (as shown in Figure \ref{fig:mealpredprob}) into binary predictions inside each window, then (ii) \emph{merge overlaps} at each global index $t$ via \emph{ratio voting}:
\[
\mathrm{ratio}(t)\;=\;\frac{|\{\text{covering windows that predict }1\text{ at }t\}|}{|\{\text{windows covering }t\}|},
\qquad
\hat y_t=\mathbbm{1}\big[\mathrm{ratio}(t)\ge \rho\big].
\]
Here, $\theta$ controls how strict a window is before merging; $\rho$ controls how strong the local consensus must be across overlapping windows. Compared with averaging probabilities on overlaps, ratio voting acts as a local majority rule that (a) is robust to isolated false positives and timing jitter, and (b) better preserves short impulses that can be over-smoothed by heavy overlap.

\paragraph{Event-level metrics.}
Let $\mathcal{T}$ (truth) and $\mathcal{P}$ (predicted) be sets of maximal $1$-segments; “overlap” means non-empty interval intersection. We report
\[
\mathrm{Recall}_{\mathrm{evt}}=
\frac{|\{\,e\in\mathcal{T}:\exists p\in\mathcal{P}\ \text{overlap}(e,p)\,\}|}{|\mathcal{T}|},\qquad
\mathrm{Precision}_{\mathrm{evt}}=
\frac{|\{\,p\in\mathcal{P}:\exists e\in\mathcal{T}\ \text{overlap}(e,p)\,\}|}{|\mathcal{P}|}.
\]

\begin{figure}[h]
    \centering
    \includegraphics[width=0.7\linewidth]{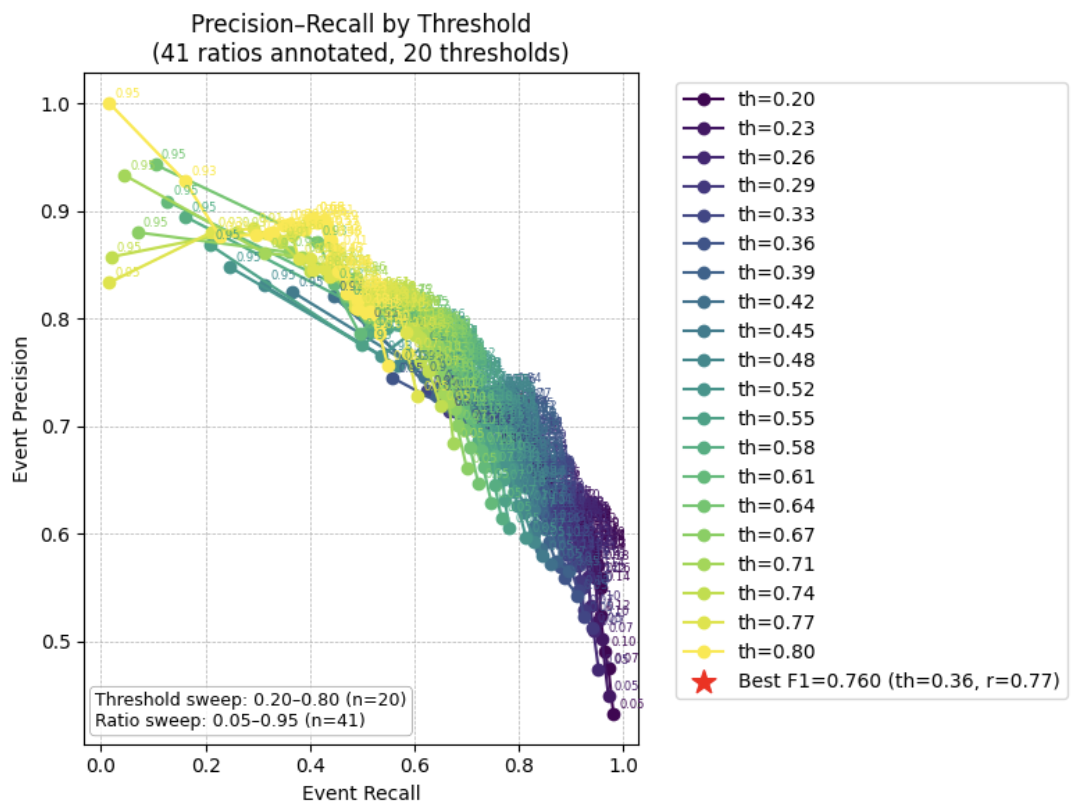}
    \caption{Validation sweep over threshold $\theta$ and ratio $\rho$.}
    \label{fig:ratio-sweep}
\end{figure}

\paragraph{Final Hyperparameters and Result}
\textbf{Selecting \boldmath$\theta$ and \boldmath$\rho$.} We sweep $(\theta,\rho)$ on the validation set to maximize the \emph{event-level} $F_{\beta}$ score ($\beta{=}1$) (As shown in Figure \ref{fig:ratio-sweep}). 
The search clarifies the trade-off: lower $\theta$ increases sensitivity within windows; higher $\rho$ demands stronger cross-window agreement to call an event. 
This joint tuning is crucial under heavy overlap ($s{=}1$), where naive averaging can under-detect brief events.

\emph{Selected hyperparameter (validation):} $\theta{=}0.36$, $\rho{=}0.77$, yielding: 

\[
\begin{aligned}
\mathrm{Recall}_{\mathrm{evt}}   &= \mathbf{80.13}\% \\
\mathrm{Precision}_{\mathrm{evt}} &= \mathbf{72.08}\%
\end{aligned}
\]

\begin{figure}[h]
\centering
\includegraphics[width=\linewidth]{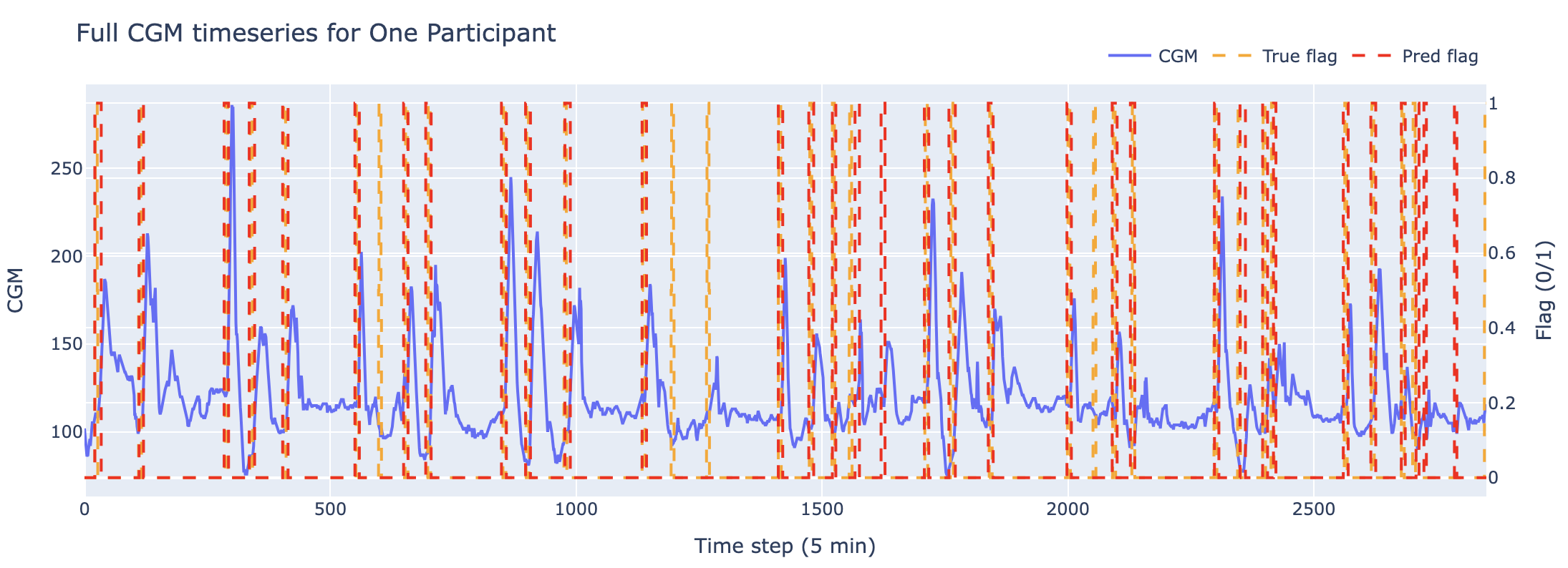}
\captionof{figure}{Full timeseries CGM meal flag predictions on one test participant after thresholding and ratio voting (CGMacros)}
\label{fig:meal-oneperson}
\end{figure}

\paragraph{Rationale for Meal Flag Deployment.}
We state that the meal flag prediction model is sufficient for deployment on AI-READI since its performance achieves a practical trade-off between precision (minimizing false positives, not labeling non-meal periods as meals) and recall (minimizing false negatives, capturing the majority of true meal periods). Specifically, the predicted meal flags align with expected physiological patterns of meal-induced glucose rises, providing a useful proxy for meal timing in a dataset where direct meal records are unavailable  (Figure \ref{fig:meal-oneperson}). 
Importantly, our aim is not to infer exact meal size or macronutrient composition, but to generate a coarse-grained signal for downstream modeling and interpretation. Given these considerations, we view the current level of performance as adequate to support as a predicted meal flag covariate in AI-READI.

\vspace{-1.0em}
\newpage
\subsection{Quick Comparisons}\label{app:quickbenchmark}

We initially compared several forecasting approaches on our CGM forecasting objective, including a persistent baseline (last value carried forward), a classical gradient boosting model (LightGBM), and neural forecasting models (DeepAR, NHITS, Seq2Seq LSTM, TFT) on chosen hyperparameters and feature engineering based on prior work \citep{ke2017lightgbm,salinas2020deepar,challu2023nhits,sutskever2014seq2seq,Lim2021TemporalFusionTransformers}. As shown in Table \ref{tab:quick_benchmark}, all methods substantially outperformed the trivial persistent baseline, confirming the importance of modeling temporal structure.

Among the neural methods, TFT achieved the best performance. In our setting—24-hour context and a 1-hour horizon with rich known-future covariates (e.g., time-of-day encodings, planned activity) and meaningful static traits (e.g., diabetes status)—TFT is particularly well aligned with the task: variable selection and attention fuse static, past, and known-future inputs, which likely explains its better accuracy. For this reason, combined with its strong accuracy, we selected TFT as the primary comparator for detailed analyses in this study and as the architecture inspiration for SSM-CGM. 

\begin{table}[ht]
\centering
\caption{Quick benchmarking across forecasting models. 
Mean absolute error (MAE; mg/dL) on AI-READI CGM data using a 24-hour context length and 1 hour horizon.}
\label{tab:quick_benchmark}
\begin{tabular}{l c}
\toprule
\textbf{Model} & \textbf{MAE (mg/dL)} \\
\midrule
Persistent (Last Value Carried Forward) & 12.30 \\
LightGBM & 9.70 \\
DeepAR & 8.70 \\
NHITS & 8.47 \\
Seq2Seq LSTM & 7.85 \\
TFT (333k params) & \textbf{7.27} \\
\bottomrule
\end{tabular}
\end{table}

\newpage

\subsection{Architecture}\label{app:arch}   

\FloatBarrier
\begin{center}
\includegraphics[width=0.92\linewidth]{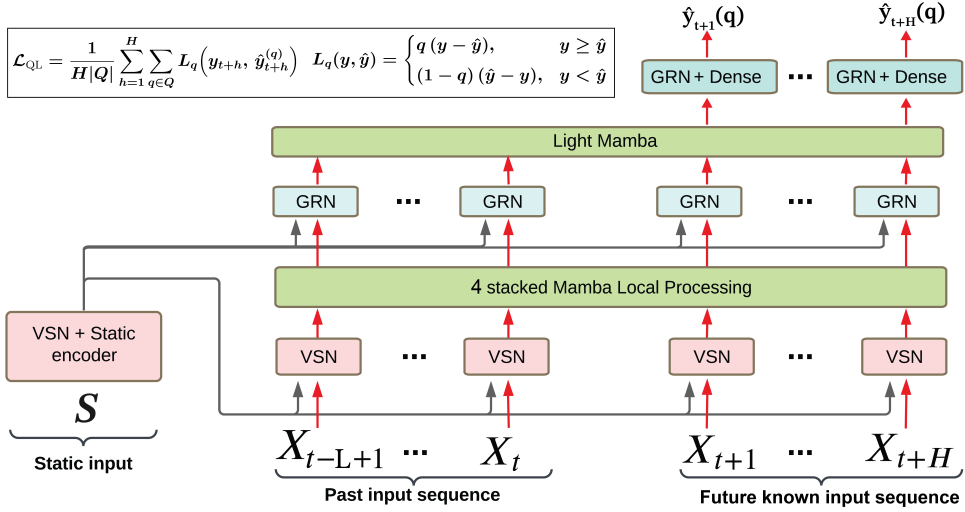}
\captionof{figure}{SSM-CGM overview: embeddings, Variable Selection Networks (static \& time-varying), SSM temporal processors (Mamba-2), static enrichment, and multi-horizon quantile head.}
\label{fig:arch-overview}
\end{center}
\FloatBarrier

\paragraph{Notation.}
At time $t$, we have historical covariates $\mathbf x^{(\mathrm{hist})}_{t-L+1:t}$, known future covariates $\mathbf x^{(\mathrm{fut})}_{t+1:t+H}$, static features $\mathbf s$, and targets $\mathbf y_{t+1:t+H}$.
Per-variable embeddings are $e_{i,t}\in\mathbb{R}^{d_x}$.

\subsection{Gated Residual Network (GRN)}
\label{app:grn}
\textbf{Intuition and purpose.}
A GRN is a compact block that \emph{stabilizes training} and \emph{controls how much nonlinearity} is injected into a representation. It combines (i) a nonlinear transform, (ii) a learnable gate deciding how much of that transform to keep, and (iii) a residual connection with normalization. This yields better gradient flow, robustness to scale/unit mismatches, and an easy way to \emph{condition} a variable by a \emph{context} (e.g., static features). In \textbf{SSM-CGM}, GRNs make variable selection and fusion smoother, more stable, and person-aware.

\textbf{Equations.}
Let $u\in\mathbb{R}^{d}$ be the block input (e.g., a per-variable embedding) and $c\in\mathbb{R}^{d_c}$ an optional conditioning vector (e.g., the static context). We write $[u;c]$ for concatenation.
\begin{align}
\hat z &= W_2\,\phi\!\bigl(W_1 [u;c] + b_1\bigr) + b_2, \label{eq:grn-core}\\
g &= \sigma\!\bigl(W_g [u;c] + b_g\bigr), \label{eq:grn-gate}\\
z &= \mathrm{LN}\!\bigl(u + g \odot \hat z \bigr). \label{eq:grn-out}
\end{align}
Here, $\phi$ is a nonlinearity (ELU/GELU), $\sigma$ is the logistic sigmoid, $\odot$ is elementwise multiplication, and $\mathrm{LN}$ is LayerNorm. 
\emph{Limiting behavior:} if $g\!\approx\!0$, the block passes $u$ through unchanged (stable linear path); if $g\!\approx\!1$, the nonlinear transform $\hat z$ is fully applied. 
If $\dim(u)\neq\dim(\hat z)$, replace $u$ by a projected skip $W_{\text{skip}}u$ in \eqref{eq:grn-out}. Dropout can be inserted after \eqref{eq:grn-core} without changing the form.

\textbf{Why it helps in \textbf{SSM-CGM}.}
(a) \emph{Context injection}: the gate $g$ learns \emph{how much} of the static context $c$ to blend into each time step; 
(b) \emph{Scale robustness}: LN + gating tame unit/scale disparities across covariates; 
(c) \emph{Regularization}: the gate can suppress unnecessary nonlinearities, reducing overfitting while preserving capacity when needed.

\subsection{Variable Selection Networks (VSN)}
\label{app:vsn}
\textbf{Intuition and purpose.}
A VSN learns, at each step, an \emph{interpretable simplex of weights} over variables: $\alpha_{i,t}\!\in\![0,1]$ with $\sum_i \alpha_{i,t}=1$. This serves two goals: (i) principled \emph{fusion} of heterogeneous covariates (different scales/reliability) into a single representation, and (ii) per-step \emph{interpretability} (“which variables did the model rely on?”). A similar VSN on \emph{static} features produces a \emph{context} $c$ that personalizes all downstream computations.

\textbf{Static VSN $\rightarrow$ context $c$.}
For static features $\{s_k\}_{k=1}^{K}$:
\begin{align}
e^{(s)}_{k} &= E^{(s)}_{k}\, s_k, \\
\tilde e^{(s)}_{k} &= \mathrm{GRN}\!\bigl([e^{(s)}_{k};\ \bar s]\bigr), \\
\beta_k &= \mathbf{v}_s^\top \tilde e^{(s)}_{k} + a_s,\qquad 
\alpha^{(s)}_{k} = \frac{\exp(\beta_k)}{\sum_{j=1}^{K} \exp(\beta_j)}, \\
c &= \sum_{k=1}^{K} \alpha^{(s)}_{k}\, P^{(s)}_{k}\, e^{(s)}_{k}.
\end{align}
$E^{(s)}_{k}$ and $P^{(s)}_{k}$ are small learned projections; $\bar s$ may be a simple summary of static variables (or omitted); and $\mathbf{v}_s\in\mathbb{R}^{d}$ and $a_s\in\mathbb{R}$ are learned parameters (scoring vector and bias) of the static VSN, with $d=\dim(\tilde e^{(s)}_{k})$. The simplex weights $\alpha^{(s)}_{k}$ reveal which static traits the model relies on to form the context $c$.

\textbf{Temporal VSN $\rightarrow$ per-step fused representation.}
For $V$ time-varying variables $\{x_{i,t}\}_{i=1}^{V}$ (CGM + time-varying covariates):
\begin{align}
e_{i,t} &= E_i\, x_{i,t}, \\
\tilde e_{i,t} &= \mathrm{GRN}\!\bigl([e_{i,t};\ c]\bigr), \\
s_{i,t} &= \mathbf{v}^\top \tilde e_{i,t} + a, \qquad
\alpha_{i,t} = \frac{\exp(s_{i,t})}{\sum_{j=1}^{V} \exp(s_{j,t})}, \\
r_t &= \sum_{i=1}^{V} \alpha_{i,t}\, P_i\, e_{i,t}. \label{eq:vsn-fused}
\end{align}

$E_i$ and $P_i$ are small learned projections; $c$ is the static context vector produced by the static VSN; and $\mathbf{v}\in\mathbb{R}^{d}$ and $a\in\mathbb{R}$ are learned parameters (scoring vector and bias) of the temporal VSN, with $d=\dim(\tilde e_{i,t})$. The simplex weights $\alpha_{i,t}$ identify, at each step, which time-varying variables the model conditionally relied on, and the fused vector $r_t$ is then fed to the temporal processor.

\textbf{Why it helps in \textbf{SSM-CGM}.}
(i) \emph{Personalization}: via $c$, per-step mixtures adapt to participant traits (age, diabetes status, site); 
(ii) \emph{Robust fusion}: softmax over GRN-conditioned scores downweights noisy or poorly scaled variables; 
(iii) \emph{Interpretability}: trajectories of $\alpha_{i,t}$ (aggregated per subject/context) indicate \emph{which} signals the model conditionally relied on (associational, not causal), supporting analysis and recommendations.

\newpage
\subsection{State-space temporal processor (Mamba)}
\label{app:mamba}
Mamba performs linear-time, streaming sequence processing via a stable state-space scan:
\begin{equation}
\label{eq:ssm-evolve}
h_{t+1} \;=\; A_t\,h_t \;+\; B_t\,x_t,
\qquad
A_t \;=\; \exp\!\big(\Delta t_t\,\Lambda_t\big),\quad \operatorname{Re}\lambda_i(\Lambda_t)\le 0,
\end{equation}
\begin{equation}
\label{eq:ssm-readout}
z_t \;=\; C_t\,h_t \;+\, D\,x_t .
\end{equation}
Here, $x_t$ denotes time-varying inputs, $h_t$ the SSM state, and $z_t$ the readout. The \emph{selective step} $\Delta t_t>0$ modulates both $A_t$ and $B_t$; $A_t$, $B_t$, and $C_t$ are light input-conditioned projections. In standard Mamba, $\Lambda_t$ is (block-)diagonal and therefore $A_t$ is (block-)diagonal; $D\,x_t$ is an optional direct/skip term (omitted in later equations). $A_t$ acts as a per-channel forget/filter; $\Delta t_t$ sets the local time scale.

\vspace{0.5em}

\textbf{Hidden attention from the SSM.}
To attribute \emph{when} the model relied on the past, we unroll the \emph{state path} (omit $D$ so the instantaneous skip at lag $0$ does not dominate). For any output index $\ell$:
\begin{align}
z_\ell 
&= \sum_{j\le \ell} K_{\ell j}\,x_j, 
\qquad 
K_{\ell j} \;=\; C_\ell \Big(\prod_{t=j}^{\ell-1} A_{\Delta t_t}\Big) B_j 
\;=\; C_\ell \exp\!\Big(\sum_{t=j}^{\ell-1}\Delta t_t\,\Lambda_t\Big) B_j. \label{eq:ssm-kernel}
\end{align}
We summarize the kernel $K_{\ell j}$ into a nonnegative scalar using the \emph{channelwise} $\ell_1$ norm,
$\psi(K_{\ell j}) \equiv \|K_{\ell j}\|_{1,\mathrm{ch}}$, and define a \emph{lag-importance} distribution with
$\varepsilon=10^{-9}$ and $\tau=1$:
\begin{align}
\alpha_{\ell j} 
\;=\; \operatorname*{softmax}_{j\le \ell}\!
\Big(\log\!\big(\psi(K_{\ell j})+10^{-9}\big)\Big). \label{eq:hidden-attn}
\end{align}
The weights $\alpha_{\ell j}$ act like a \emph{hidden attention} over past lags $j$ for each forecast index $\ell$.

\vspace{0.75em}
\newpage
\subsection{Mamba-2 MES-style}
\label{app:mamba2}

\textbf{Motivation.}
Mamba-2 factorizes channels into multiple heads with grouped states and an explicit $(A,B,C,D)$ parameterization, improving throughput and long-range capture. To make head-wise attributions \emph{comparable}, we \emph{share} the value space and the readout across heads, analogous in spirit to the \emph{interpretable multi-head self-attention} of TFT where common value projections ease cross-head comparison.

\textbf{Per-head SSM and shared projections.}
Let $H$ be the number of heads. We introduce a shared value projection $V:\mathbb{R}^{d_x}\!\to X$ and a shared readout $C_{\mathrm{sh}}:\mathbb{R}^{d_h}\!\to \mathbb{R}^{d_z}$. Each head $h$ has its own stable dynamics and expansion:
\begin{align}
v_t &= V\,x_t, \qquad v_t\in X, \label{eq:m2-value}\\
h^{(h)}_{t+1} &= A^{(h)}_{\Delta t_t}\,h^{(h)}_t + B^{(h)}\,v_t, \qquad A^{(h)}_{\Delta t_t}=\exp\!\big(\Delta t_t\,\Lambda^{(h)}_t\big), \label{eq:m2-evolve}\\
z^{(h)}_t &= C_{\mathrm{sh}}\,h^{(h)}_t \;+\; D\,x_t, \qquad z_t=\sum_{h=1}^{H} z^{(h)}_t. \label{eq:m2-readout}
\end{align}
The skip $D$ is shared (or omitted). Unrolling the \emph{state path} yields the head-wise kernel
\begin{align}
K^{(h)}_{\ell j} 
\;=\; C_{\mathrm{sh}}\!\left(\prod_{t=j}^{\ell-1} A^{(h)}_{\Delta t_t}\right) B^{(h)}
\;=\; C_{\mathrm{sh}}\exp\!\Big(\sum_{t=j}^{\ell-1}\Delta t_t\,\Lambda^{(h)}_t\Big) B^{(h)}, \label{eq:m2-kernel}
\end{align}
and the output decomposition
\begin{align}
z_\ell \;=\; \sum_{h=1}^{H}\sum_{j\le \ell} K^{(h)}_{\ell j}\,v_j. \label{eq:m2-output}
\end{align}

\textbf{Shared value projection $V$.}
We tested several ways to define $V$: (i) a squeeze-and-excitation (SE) mechanism across heads—first averaging over the SSM channel dimension $P$ to obtain $s_t\!\in\!\mathbb{R}^{H}$, then applying a small MLP $H\!\to\!H/r\!\to\!H$ and a softmax over $H$ to weight and aggregate heads; (ii) head-wise max pooling; and (iii) a simple head-wise mean. We ultimately adopt the head-wise mean as the shared value because it is simple, stable, and effective:
\[
v_t \;=\; \tfrac{1}{H}\sum_{h=1}^{H} x_t^{(h)} \in \mathbb{R}^{P}.
\]
This choice is a pragmatic baseline, and richer value-sharing schemes (e.g., learned head gating or data-dependent projections) are left for future work.

\textbf{Head-wise hidden attention and comparability.}
Because \emph{both} the value projection $V$ and the readout $C_{\mathrm{sh}}$ are shared, the scalar summaries $\psi\!\big(K^{(h)}_{\ell j}\big)$ live on a common scale across heads. We define per-head lag-weights

\begin{align}
\alpha^{(h)}_{\ell j} 
\;=\; \operatorname*{softmax}_{j\le \ell}\!
\left(\frac{\log\!\big(\psi(K^{(h)}_{\ell j})+\varepsilon\big)}{\tau}\right), \label{eq:m2-alpha}
\end{align}
and, if desired, a head-aggregated map
\begin{align}
\bar\alpha_{\ell j} \;=\; \operatorname*{softmax}_{j\le \ell}\!\left(\frac{1}{\tau}\log\!\Big(\tfrac{1}{H}\sum_{h=1}^{H}\psi(K^{(h)}_{\ell j})+\varepsilon\Big)\right). \label{eq:m2-agg}
\end{align}
This mirrors TFT’s interpretable multi-head attention: by fixing the \emph{value} ($V$) and the \emph{readout} ($C_{\mathrm{sh}}$) across heads, differences in $\alpha^{(h)}_{\ell j}$ reflect \emph{temporal dynamics} (via $\Lambda^{(h)}_t,B^{(h)}$), not arbitrary head-specific value/readout scalings.

\textbf{Practical notes.}
We parameterize each $\Lambda^{(h)}_t$ to have nonpositive real parts (e.g., diagonal with $\Lambda^{(h)}_{kk}=-\exp(\theta^{(h)}_{kk})$) to ensure stability; $\tau$ tunes sharpness of the attribution maps; $\psi$ used is the channelwise $\ell_1$ norm. Hidden-attention maps (\ref{eq:hidden-attn}, \ref{eq:m2-alpha}, \ref{eq:m2-agg}) are \emph{associational} and complement VSN importances: together they answer \emph{what} (variables) and \emph{when} (lags) the model relied on for CGM forecasting.

\newpage
\subsection{Benchmarking protocol, fairness, and reproducibility}
\label{app:benchmark}

\paragraph{Goal.}
We compared \textbf{SSM\textendash CGM} against a strong baseline (Temporal Fusion Transformer, \textbf{TFT}) under \emph{matched budget}, \emph{identical data/views}, and \emph{aligned hyperparameter axes}. Any performance difference should stem solely from the temporal processor (SSM/Mamba vs.\ LSTM{+}attention).

\paragraph{Setup and parity.}

We use the same AI\textendash READI cohort, features, and splits (Sec.~\ref{sec:dp}) with identical preprocessing. The task is a multi\textendash horizon forecast ($H{=}12$ steps at 5\,min) with the same covariate interface (static / time\textendash varying historical / time\textendash varying future) and context lengths of 12\,h, 24\,h, and 48\,h. \textbf{TFT} (via \texttt{pytorch\textendash forecasting}) uses VSNs, LSTMs, interpretable multi\textendash head attention, and a quantile head; \textbf{SSM\textendash CGM} keeps the exact same interface/head and replaces only the temporal stack with Mamba\textendash 2 (MES\textendash style). Both models optimize the same QuantileLoss ($q\in\{0.1,0.5,0.9\}$) with Adam, early stopping (patience 15), a 30\textendash epoch cap, and identical validation cadence(5 times per epoch) /batching/teacher forcing; we report QuantileLoss, MAE, and RMSE (Table ~\ref{tab:bench-mini}).

\begin{table}[h]
\centering
\small 
\setlength{\tabcolsep}{6pt} 
\renewcommand{\arraystretch}{1.1} 
\caption{Best hyperparameters used in all reported results.}
\label{tab:best-hparams}
\begin{tabular}{lcc}
\toprule
\textbf{Setting} & \textbf{TFT (baseline)} & \textbf{SSM--CGM (Mamba--2, MES)} \\
\midrule
\multicolumn{3}{c}{\hspace{+7em}\it Architecture} \\
\midrule
Hidden size $d_{\text{model}}$ & 128 & 128 \\
Attention heads & 8 & 8 (expand=4, headdim=64) \\
Temporal depth & LSTM layers=4 & Mamba depth=4 \\
Dropout & 0.2 & 0.2 \\
$d_{\text{state}}$ & --- & 128 \\
$d_{\text{conv}}$ & --- & 8 \\
Expand factor & --- & 4 \\
Head dimension & --- & 64 \\
Groups (\texttt{ngroups}) & --- & 1 \\
Parameter count & 1,793,376 & 1,762,714 \\
\midrule
\multicolumn{3}{c}{\hspace{+7em}\it Training} \\
\midrule
Loss & \multicolumn{2}{c}{\hspace{-2em}QuantileLoss $q \in \{0.1,0.5,0.9\}$} \\
Learning rate & $1\times10^{-3}$ & $1\times10^{-3}$ \\
Batch size & 32 & 32 \\
Max epochs & 30 & 30 \\
Early stopping patience & 15 & 15 \\
Validation check interval & 0.2 (per epoch) & 0.2 (per epoch) \\
\midrule
\multicolumn{3}{c}{\hspace{+7em}\it Runtime / system} \\
\midrule
Hardware & \multicolumn{2}{c}{\hspace{-2em}4$\times$A100 (DDP), TF32 enabled} \\
Chunk size & --- & 128 \\
$dt$ limit & --- & $(10^{-3},\,10)$ \\
Learnable init states & --- & True \\
\bottomrule
\end{tabular}
\end{table}

\paragraph{Hyperparameter search.}
Because \textbf{SSM\textendash CGM} is designed to match the \textbf{TFT} interface, we use the \emph{same axes and ranges} for both models (e.g., LSTM depth for TFT mirrors Mamba depth for SSM). We run an \emph{random search} with \emph{$\sim$25 configurations per model}, sampling uniformly over the shared space below. Remaining Mamba\textendash specific settings are fixed based on prior Mamba literature (see Table ~\ref{tab:best-hparams}):
\[
\begin{aligned}
&\text{Hidden size } d_{\text{model}} \in \{64, 128\}, \\
&\text{Heads (TFT) / head\textendash dim config (SSM)} \in \{2, 4, 8\}, \\
&\text{Depth} \in \{1, 2, 4\} \quad (\text{LSTM layers for TFT; Mamba depth for SSM}), \\
&\text{Learning rate} \in \{\,3\!\times\!10^{-3},\,1\!\times\!10^{-3}\}, \\
&\text{Batch size} \in \{32, 64\}. \\
\end{aligned}
\]
The primary selection criterion is validation QuantileLoss; ties break by MAE, then RMSE, then parameter count. The best configuration found on the 24\,h context has been transferred to 48\,h  and 12\,h for both models. Notably, the best settings are \emph{identical} for TFT and SSM\textendash CGM (Table ~\ref{tab:best-hparams}), consistent with the aligned axes and shared interface (that makes similar number of params: 2\% less for SSM CGM).

Runs use 4$\times$A100\,40GB (DDP) on Google Cloud; TF32/mixed precision follow vendor defaults; checkpoints are saved every epoch.
On the 48\,h context, training time per epoch was approximately 1.5 hours on these 4 GPU for both SSM\textendash CGM and TFT.

\paragraph{Limitations.}
We transfer hyperparameter to 12\,h and 48\,h context without a dedicated retuning to control compute. This could under\textendash optimize either model for others context length. We mitigate this by enforcing strict budget symmetry and aligned axes. A targeted 48\,h (and 12\,h) refinement is left for future work.

\newpage
\subsection{VSN Interpretability }\label{app:interp-vsn}

\begin{figure}[h]
    \centering
    \includegraphics[width=0.49\linewidth]{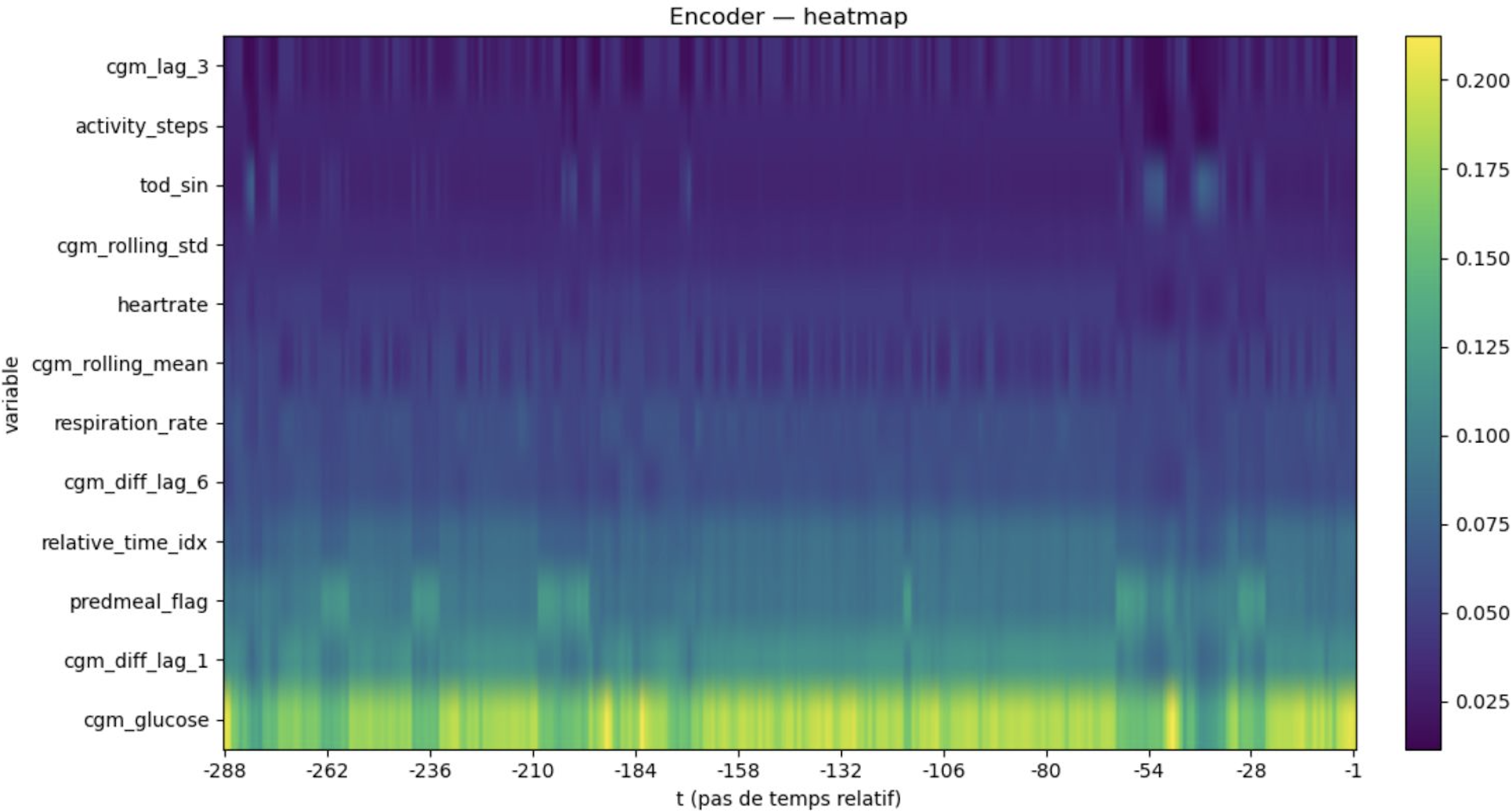}
    \hfill
    \includegraphics[width=0.49\linewidth]{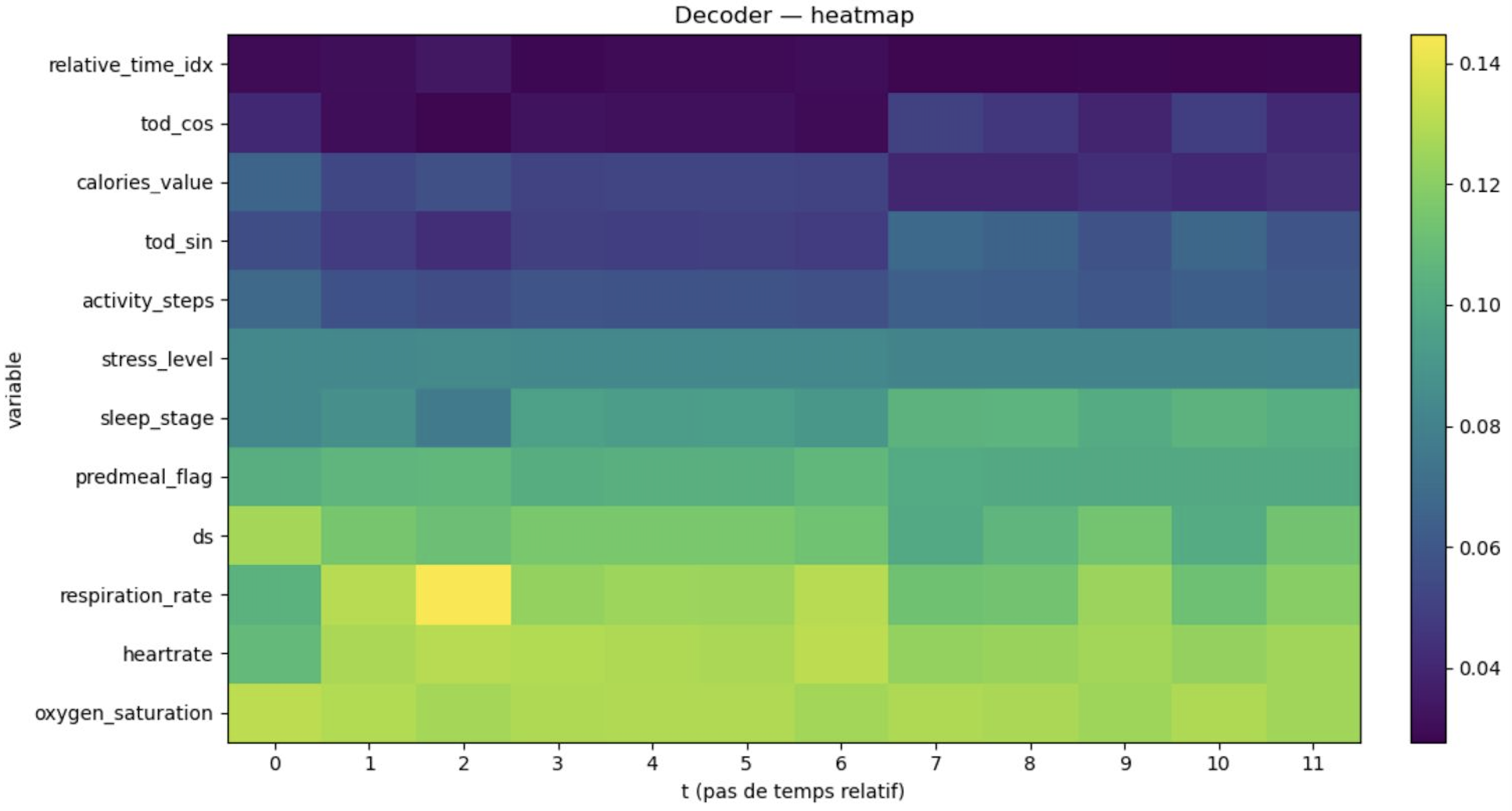}
    \captionof{figure}{VSN variable by time importance for the encoder and the decoder}
    \label{fig:VSNtemporalimp}
\end{figure}

These VSN results indicate that wearable covariates are informative in both the encoder and the decoder. Their time-varying weights reveal when the model conditionally relies on each signal to form its forecast. In the individual example shown in \Cref{fig:VSNtemporalimp}, we observe substantial within-subject variability over time in both stages. For instance, peaks in the weight assigned to the predicted-meal flag within the encoder align with when meal flags are present, indicating that when a meal is likely, the model allocates more mass to this covariate and correspondingly shifts its forecast.

\newpage
\subsection{Temporal attention Interpretability}\label{app:interp-hidden}
\vspace{-1.0em}
\begin{center}
  \centering

  \newcommand{\imgdir}{figures/interpretability/appendix_interpretability/Mean_stacked}
  \newcommand{\avgdir}{figures/interpretability/appendix_interpretability/meanmeanstacked}

  \setlength{\tabcolsep}{2pt}
  \renewcommand{\arraystretch}{0.8}
  \newcommand{\tilewd}{0.15\linewidth}

  \newcommand{\headrow}[1]{%
    \textbf{Head~#1} &
    \includegraphics[width=\tilewd]{\imgdir/mean_SA_l_0_h_#1.png} &
    \includegraphics[width=\tilewd]{\imgdir/mean_SA_l_1_h_#1.png} &
    \includegraphics[width=\tilewd]{\imgdir/mean_SA_l_2_h_#1.png} &
    \includegraphics[width=\tilewd]{\imgdir/mean_SA_l_3_h_#1.png} \\
  }

  \newcommand{\meanrow}{%
    \textbf{Layer avg} &
    \includegraphics[width=\tilewd]{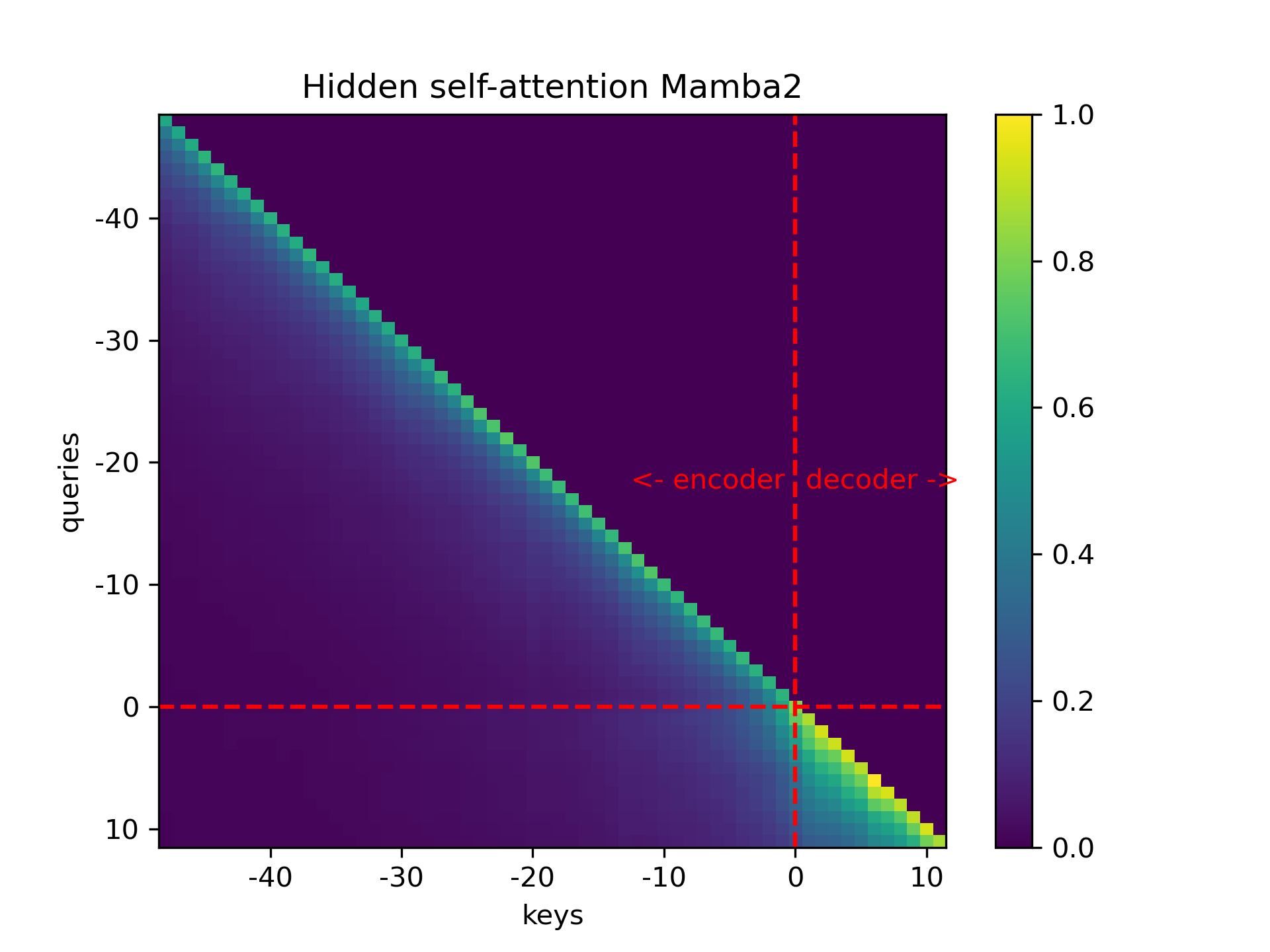} &
    \includegraphics[width=\tilewd]{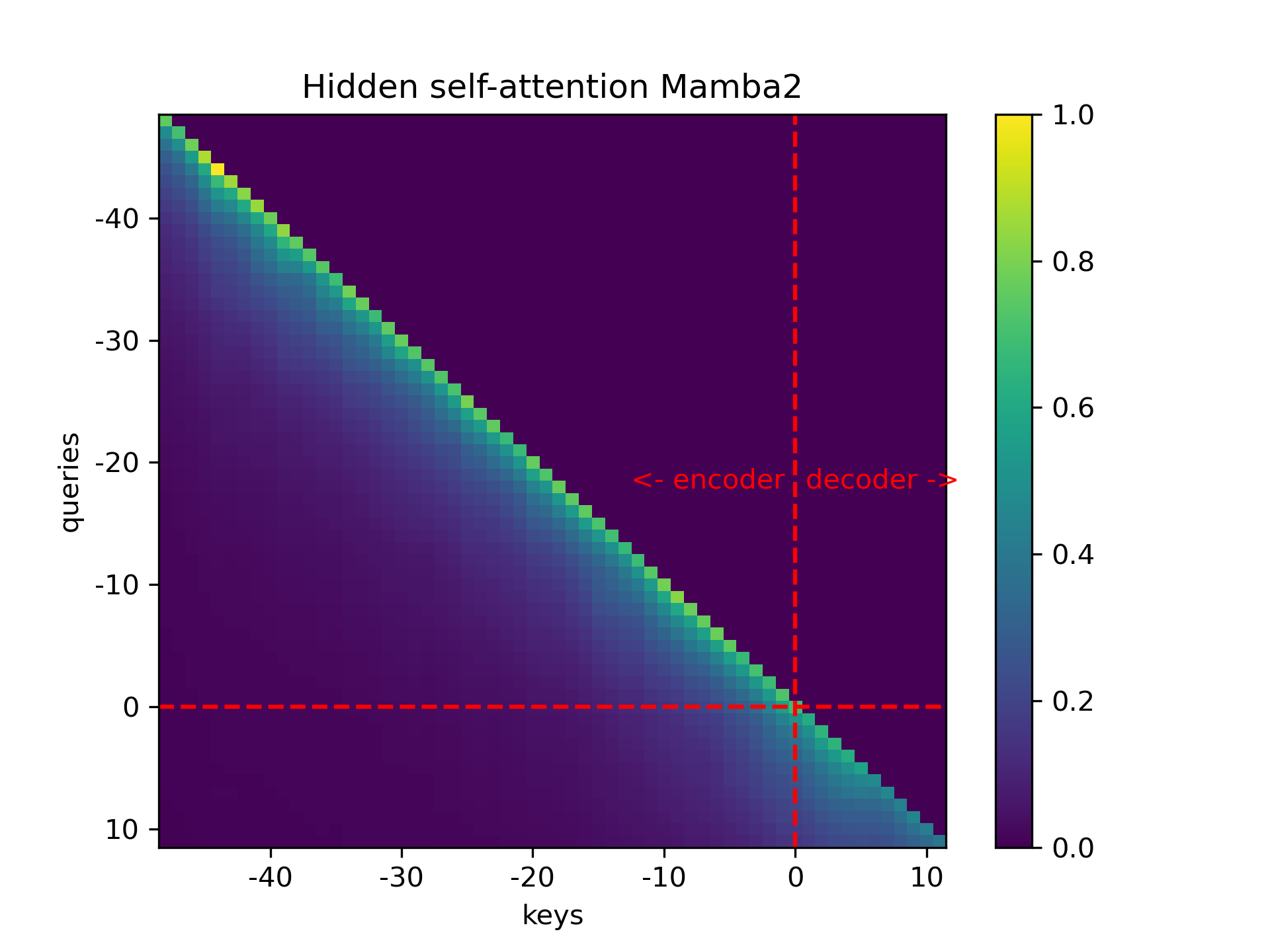} &
    \includegraphics[width=\tilewd]{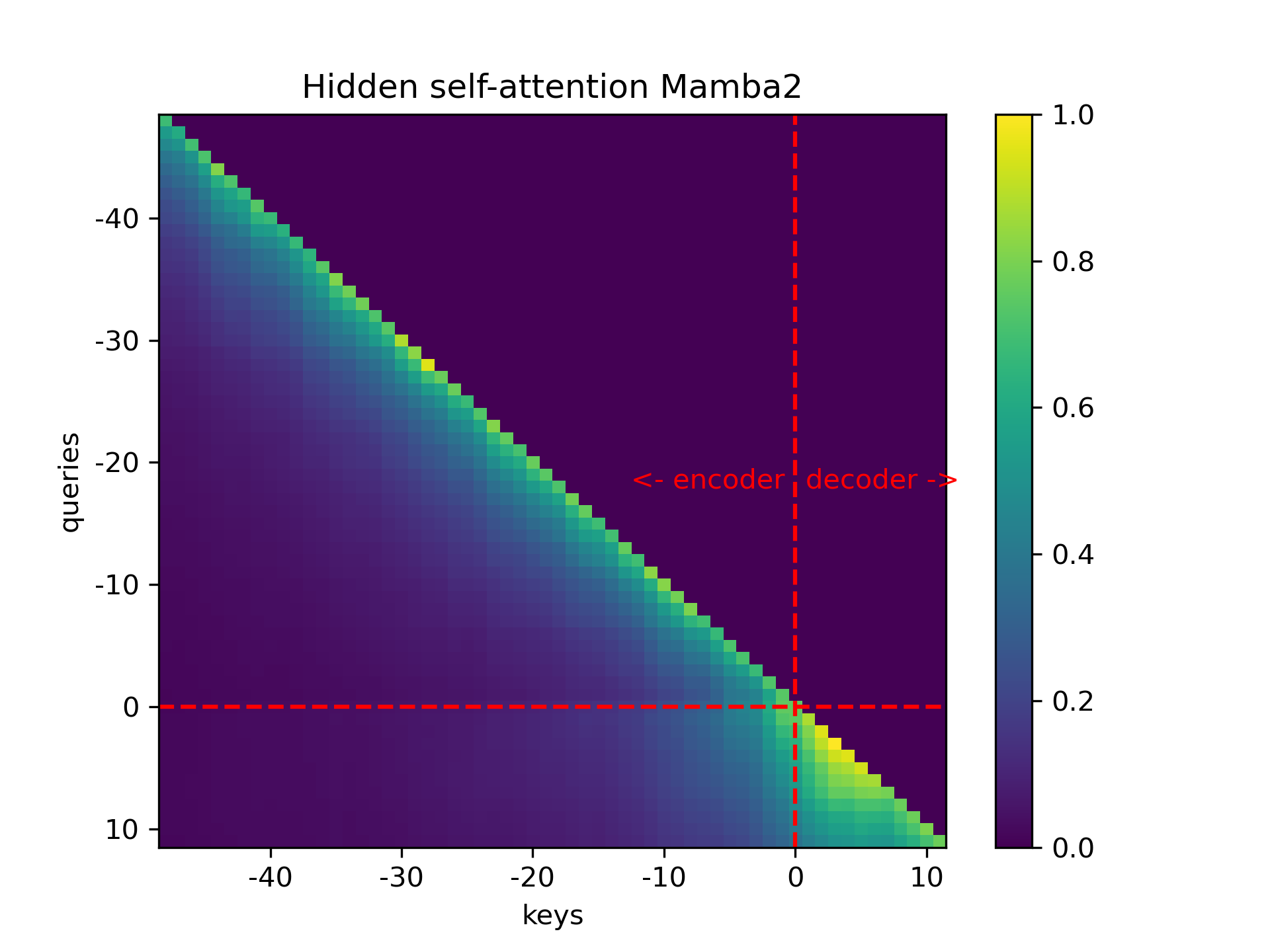} &
    \includegraphics[width=\tilewd]{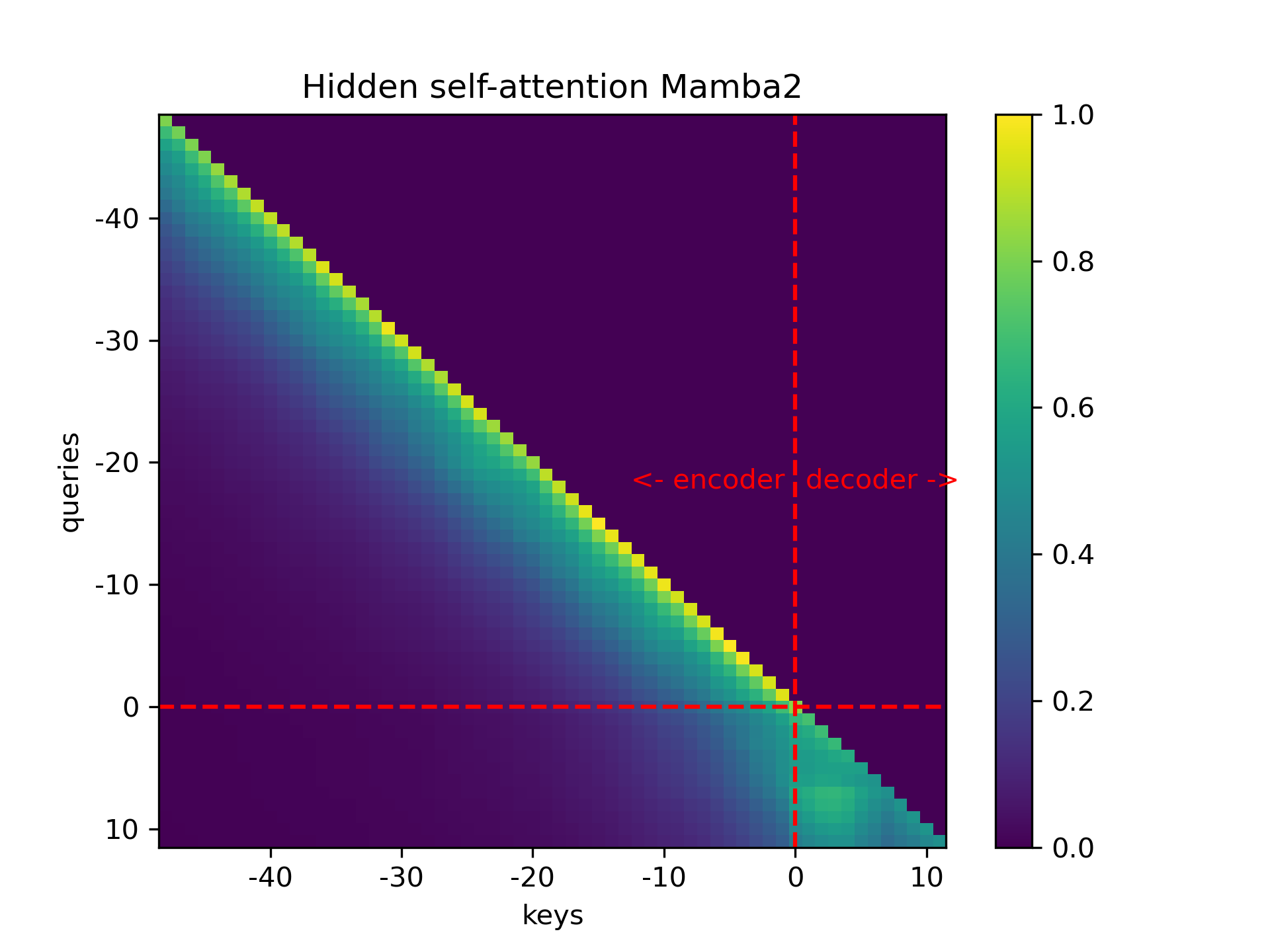} \\
  }

  \resizebox{\textwidth}{!}{\begin{tabular}{@{}>{\raggedleft\arraybackslash}p{16mm}*{4}{c}@{}}
    & \textbf{Layer 0} & \textbf{Layer 1} & \textbf{Layer 2} & \textbf{Layer 3} \\
    \headrow{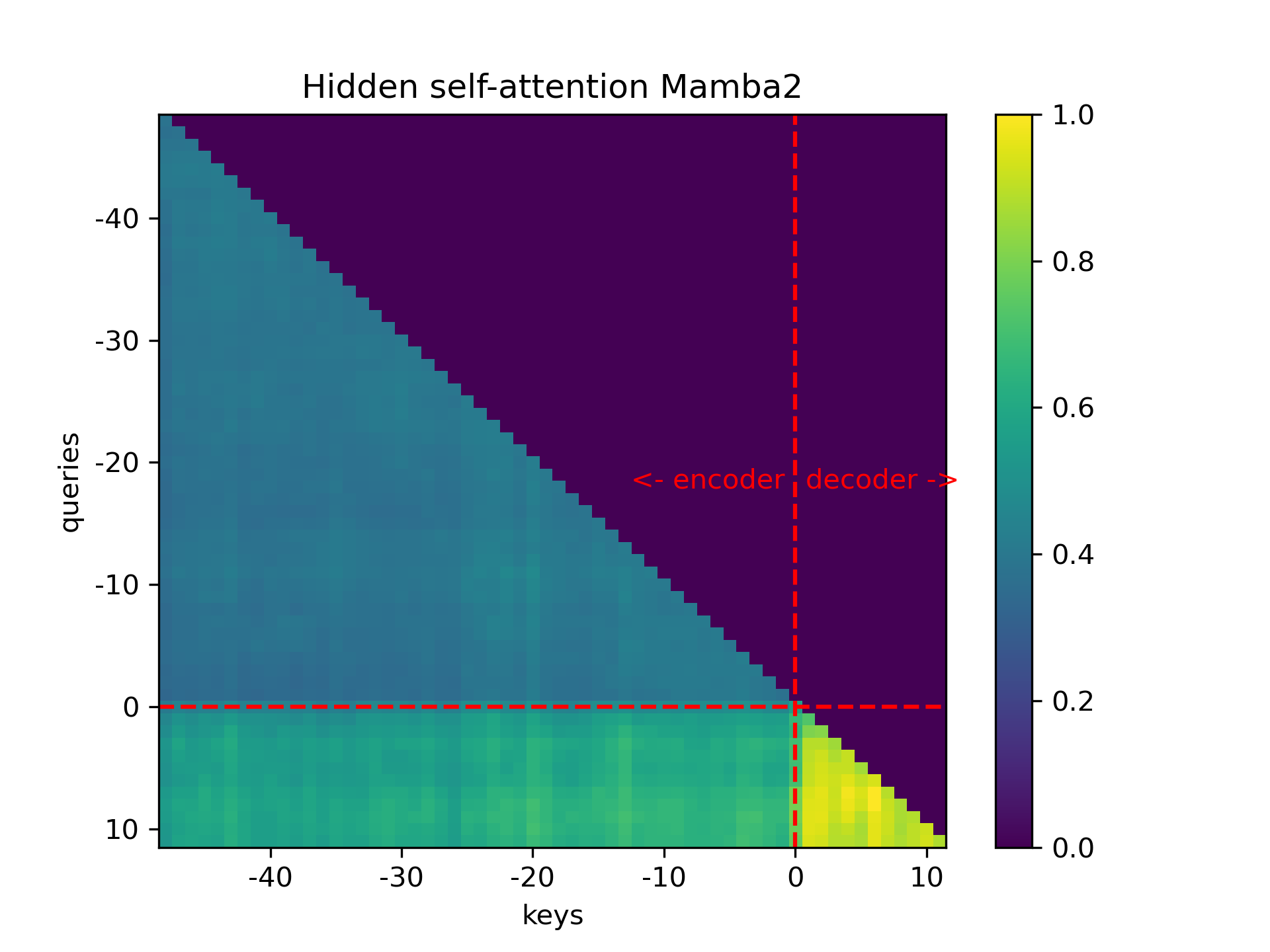}
    \headrow{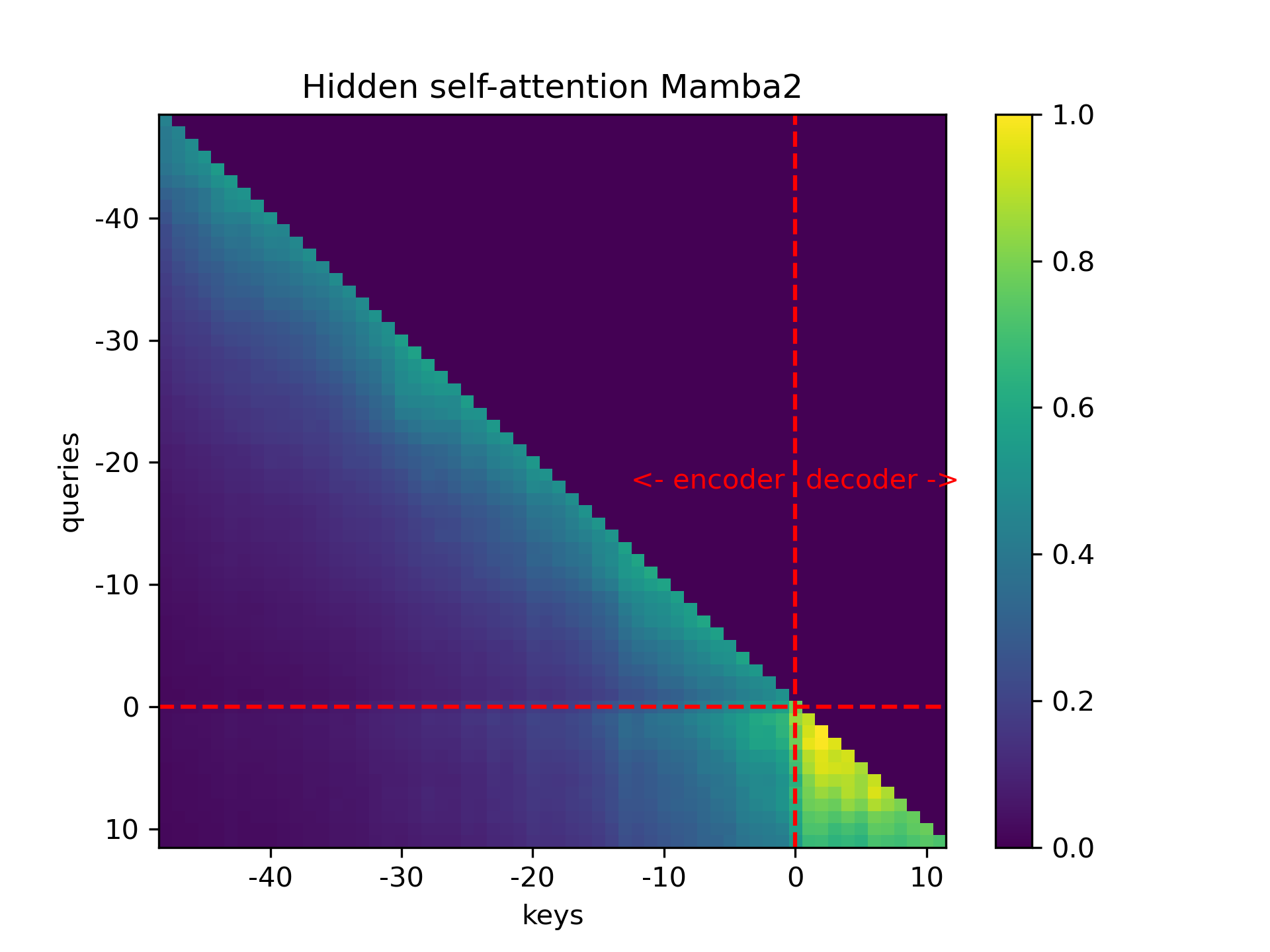}
    \headrow{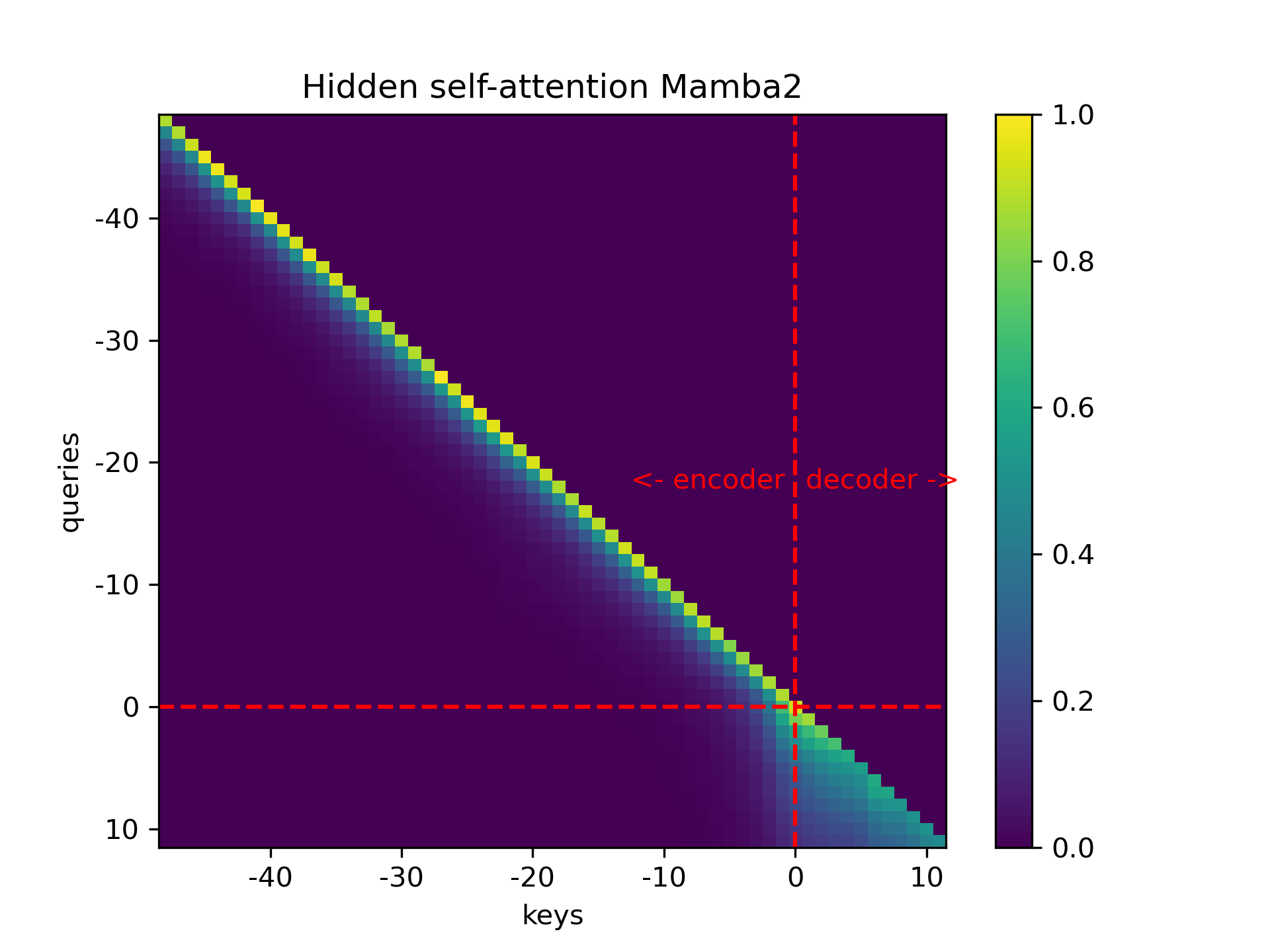}
    \headrow{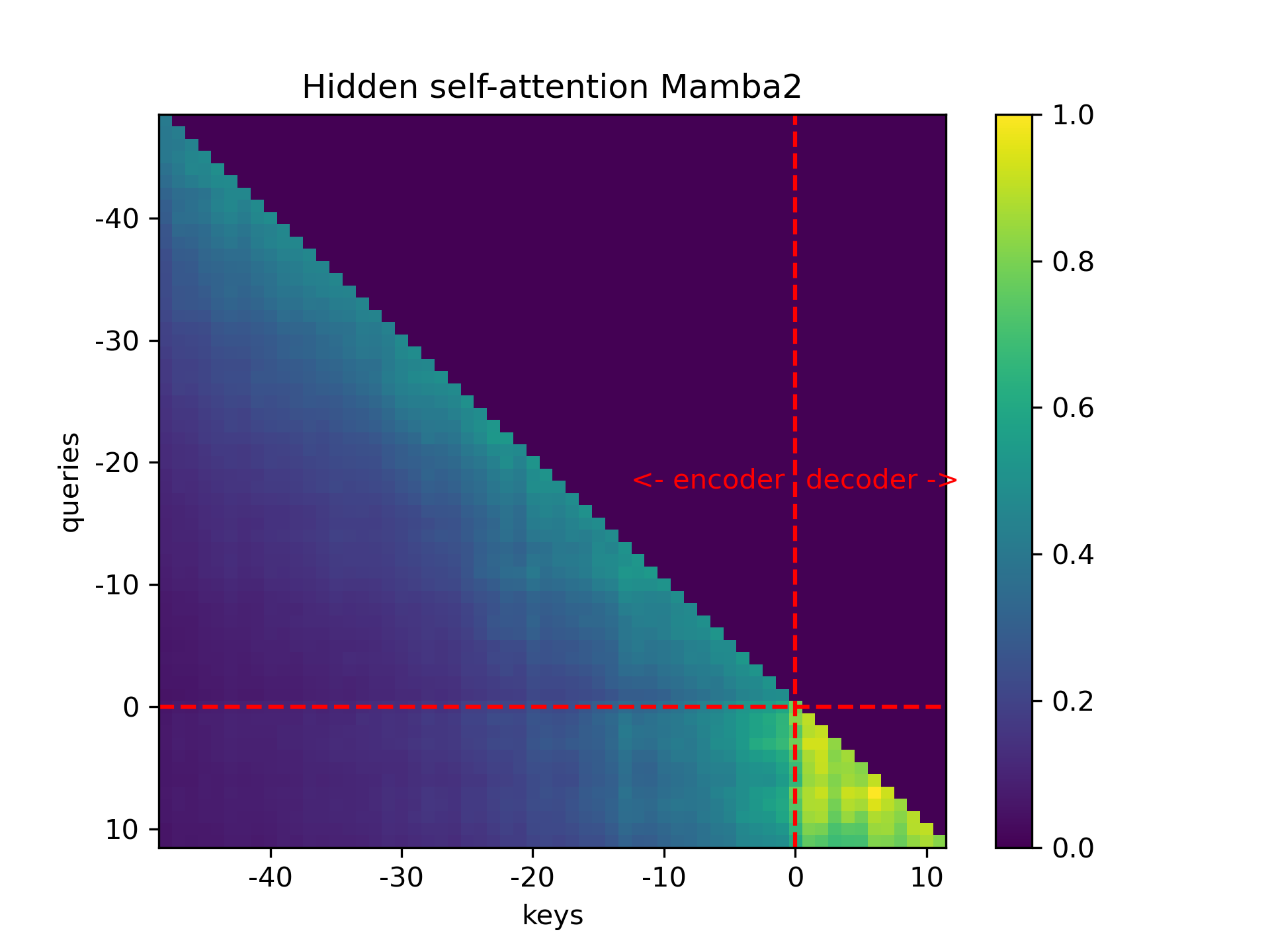}
    \headrow{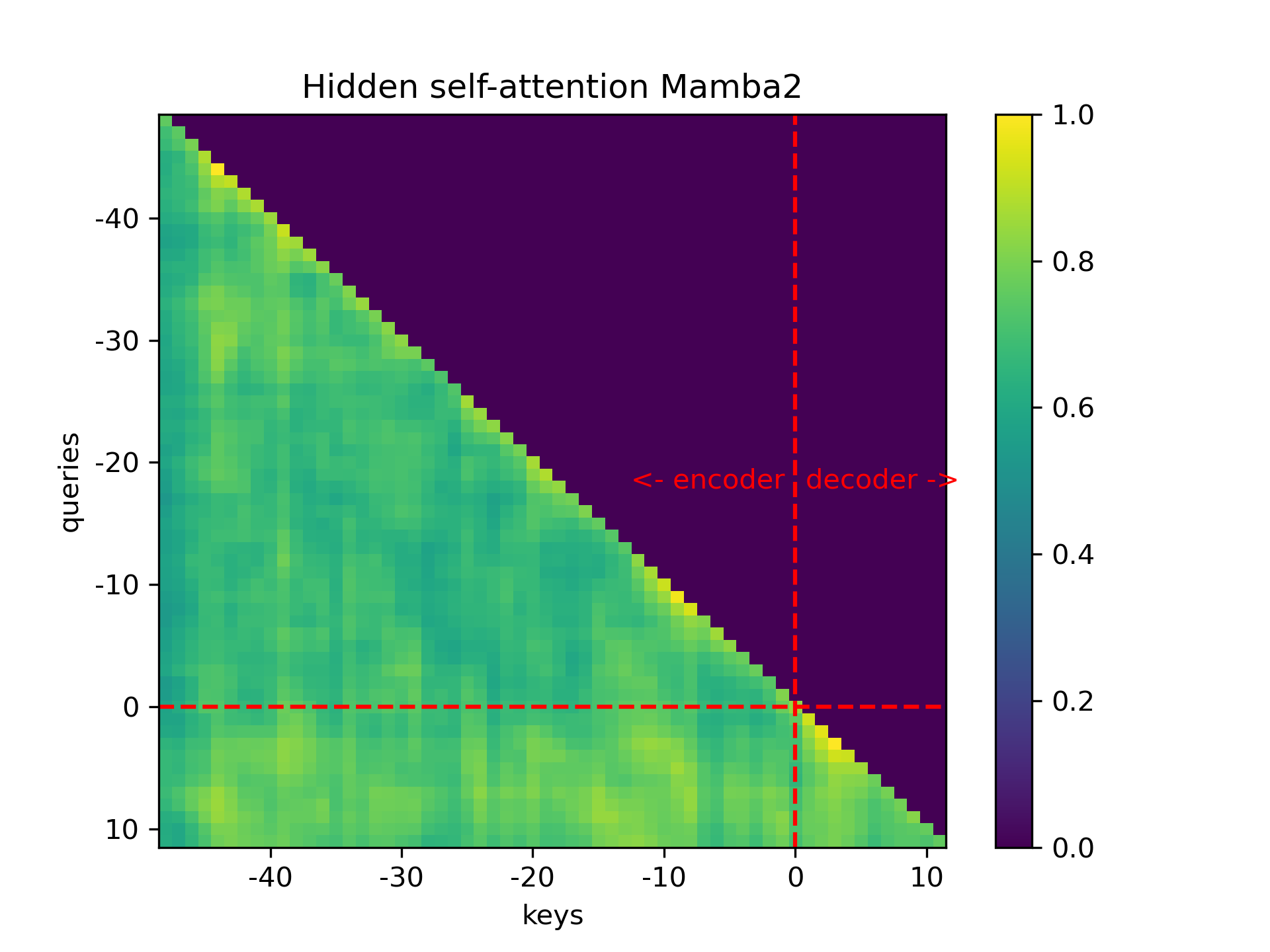}
    \headrow{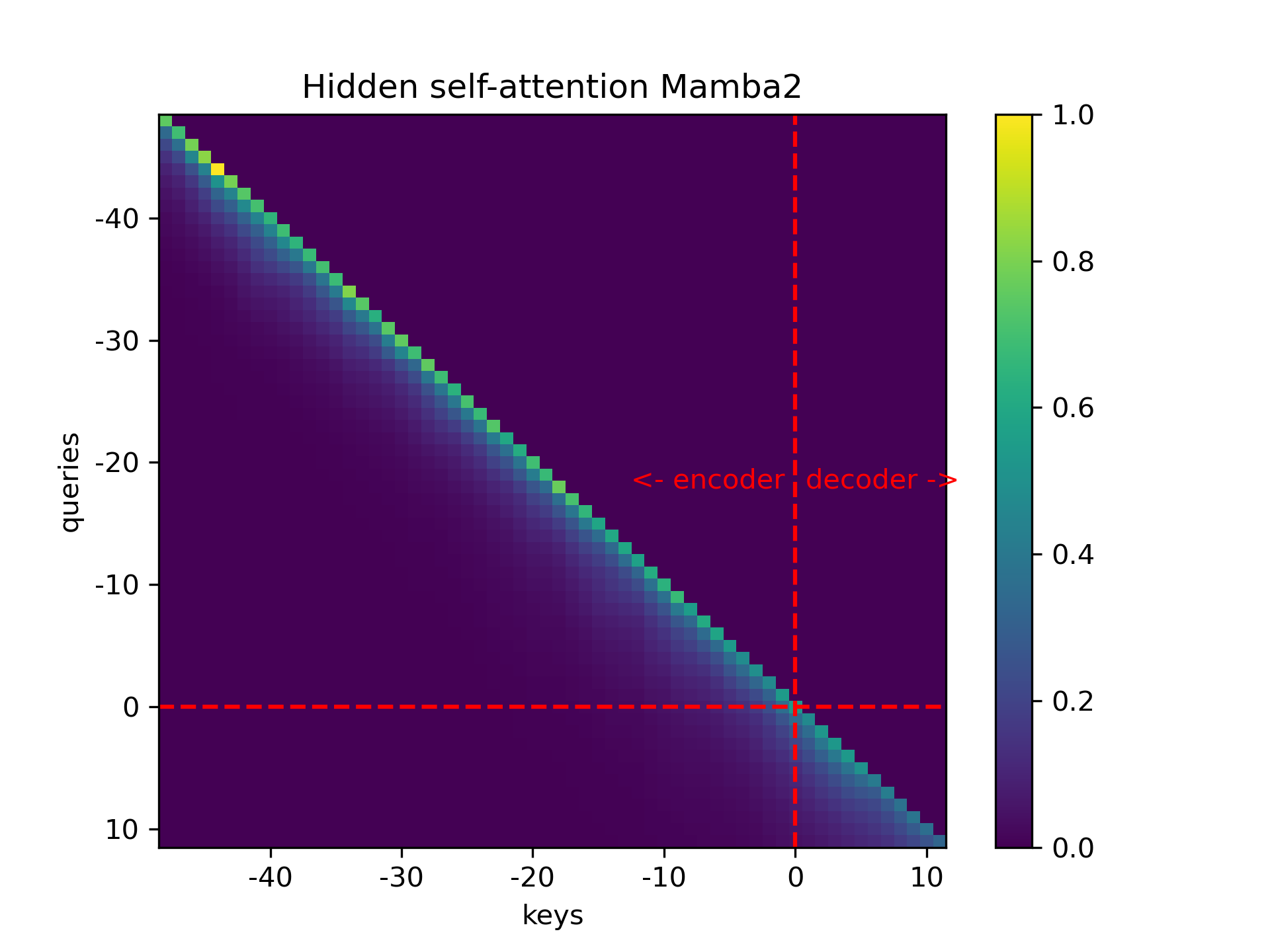}
    \headrow{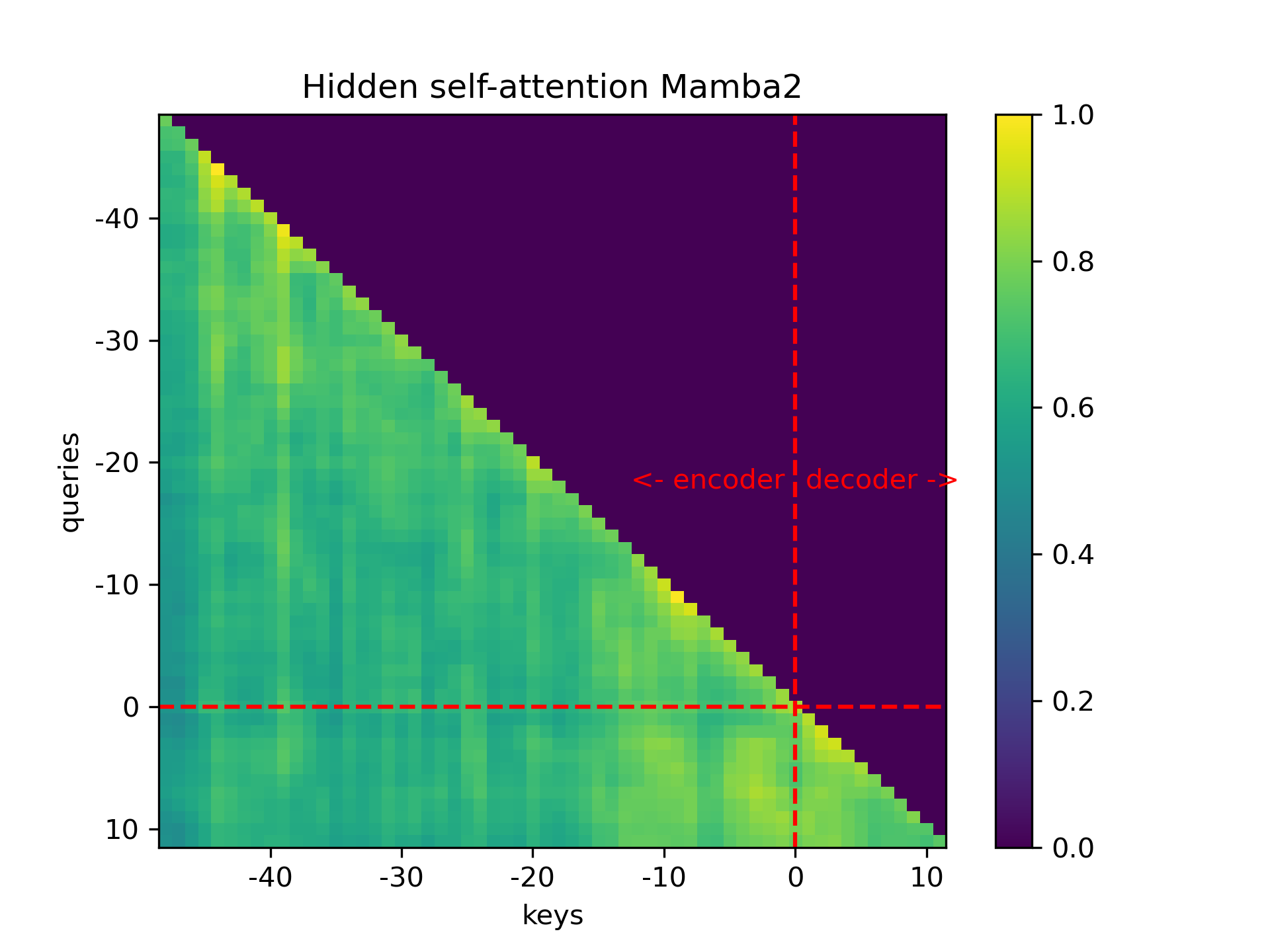}
    \headrow{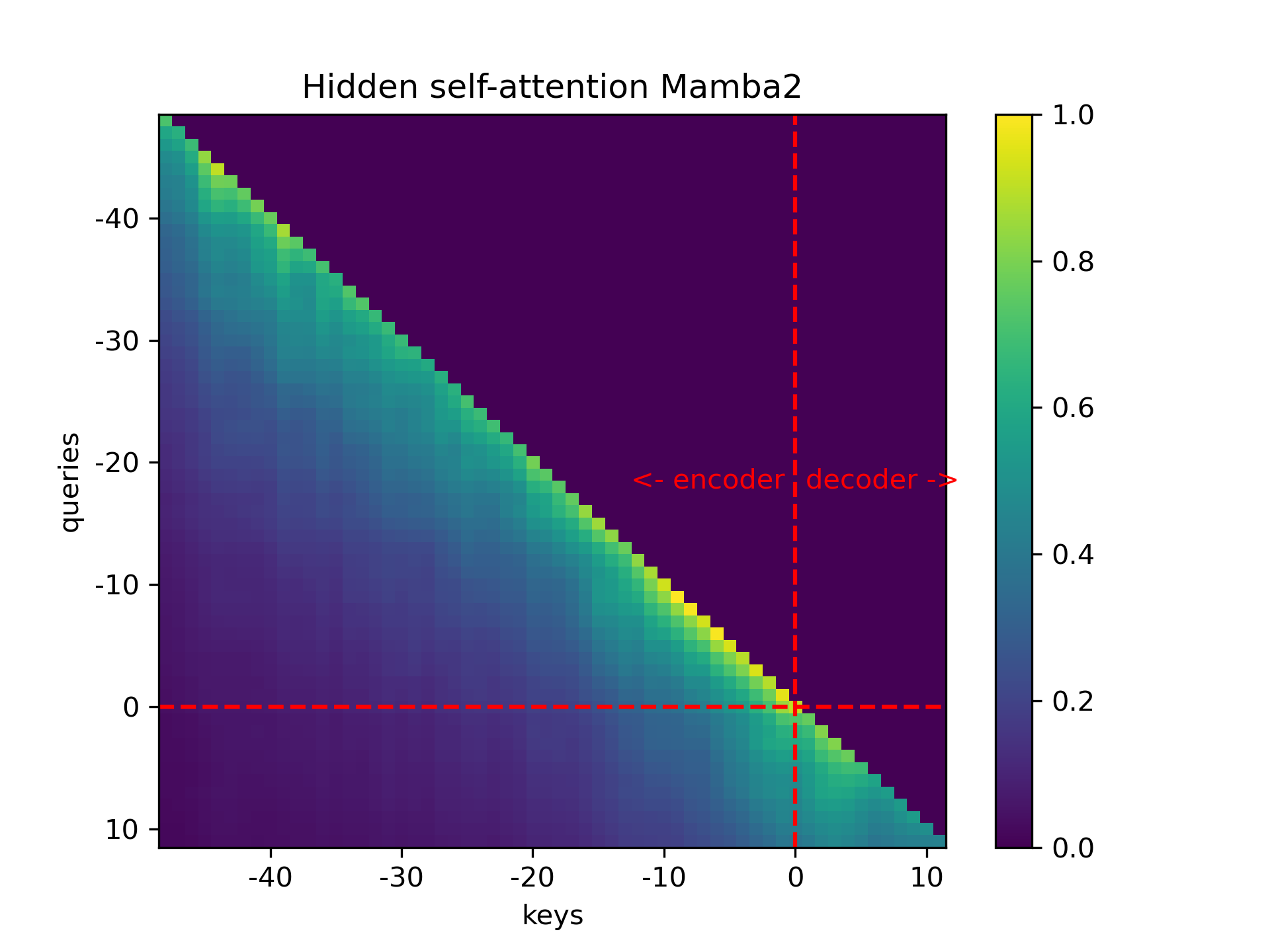}
    \meanrow
  \end{tabular}}
  \captionsetup{type=figure}
  \caption{Average hidden self-attention maps over the last 60 time steps across all individuals, by Mamba head and layer depth (stacked Mamba).}
  \label{fig:hidden-attn-layer-head-grid}
\end{center}

Hidden-attention maps derived from the SSM unrolling across all Mamba layers provide a fine-grained view of \emph{when} past information is used for each prediction step. At the individual level, these maps expose specific lag structures; at the cohort level, layer- and head-wise averages clarify complementary roles across the stack.

Analysis of the maps in Figure~\ref{fig:hidden-attn-layer-head-grid} gives intuition on the division of temporal labor across the stacked Mamba: earlier layers appear to emphasize very recent lags that stabilize short-horizon decoder steps, whereas deeper layers tend to place more mass further back in the encoder window, consistent with slower dynamics and circadian structure, before passing representations to the light-Mamba. This pattern is reflected by head-wise maps that partition the temporal context within each layer and by layer-average rows showing a progressive broadening of the effective receptive field.

\begin{figure}[h]
    \centering
    \includegraphics[width=0.5\linewidth]{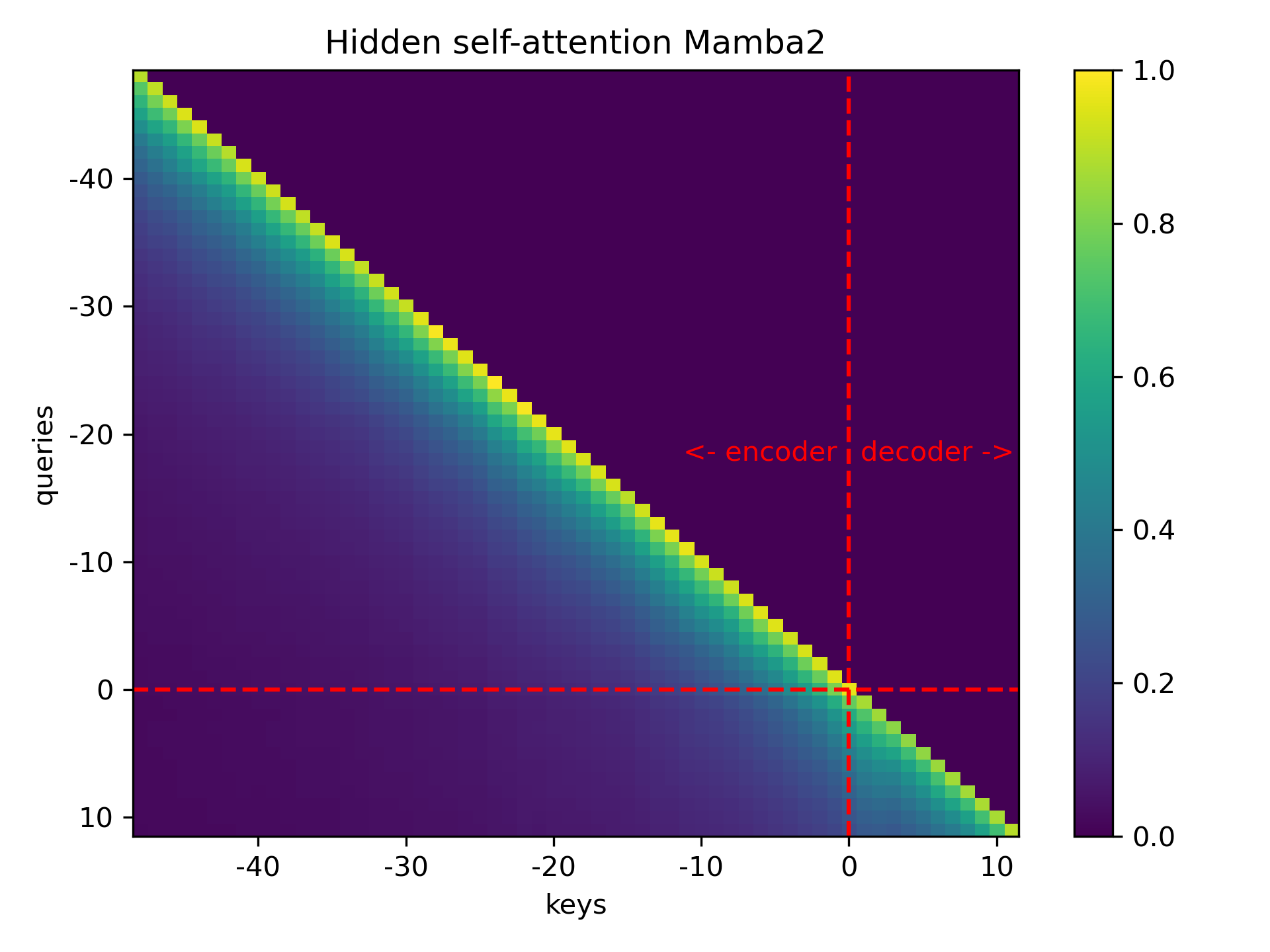}
    \captionof{figure}{Average hidden self-attention, over the last 60 time steps, across all participants (light Mamba).}
    \label{fig:HiddenAttnAvg}
\end{figure}

Figure~\ref{fig:interpretable}C displays the light-Mamba self–hidden-attention over the last 60 time steps. Comparing the cohort-average map with individual maps (Figure \ref{fig:HiddenAttnAvg}) shows that certain past segments are consistently retained and influence the decoder, while others are rapidly down-weighted. This pattern is consistent with CGM dynamics in which very recent history dominates, punctuated by salient events that remain influential over longer lags.

\clearpage
\newpage
\newpage
\subsection{Interpretability - Static Covariates}\label{app:interp-static}

Static features in SSM-CGM propagate information across sequential inputs through a static variable selection network (VSN) combined with a static encoder that handles both categorical and continuous features (Figure \ref{fig:ModArch}). The static VSN enables computation of feature importances for each static feature (Table \ref{tab:static_var_importance}). Participant ID had the highest feature importance, indicating that individual variation in glucose dynamics is important to encode for forecasting CGM. The next most important features were  variance in the glucose levels (glucose scale) and diabetes status suggesting that capturing differences in participants with high vs low glycemic variability improves forecasting. Other static features such as age, and clinical site also show importance, though to a lesser extent.

\begin{table}[ht]
\centering
\caption{Static Variable Importances}
\label{tab:static_var_importance}
\begin{tabular}{l r}
\hline
\textbf{Variable} & \textbf{Importance} \\
\hline
participant id      & 0.4707 \\
glucose scale  & 0.2553 \\
diabetes status         & 0.0921 \\
encoder length      & 0.0729 \\
age                  & 0.0439 \\
clinical site       & 0.0338 \\
glucose center & 0.0313 \\
\hline
\end{tabular}
\end{table}

\begin{figure}[ht]
    \centering 
    \includegraphics[width=0.8\textwidth]{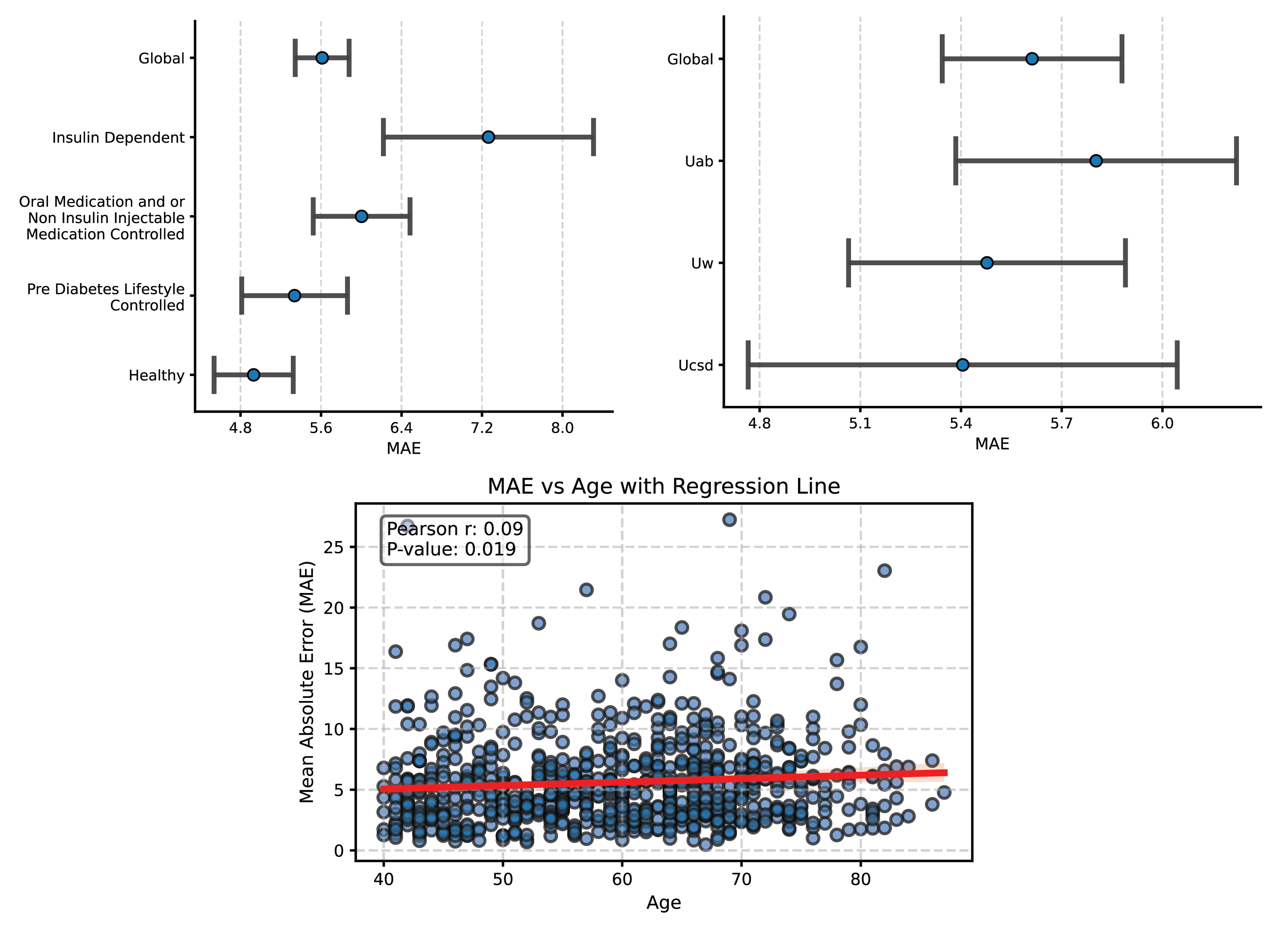}
    \caption{MAE error by category in static features.}
    \label{fig:StaticErrors}
\end{figure}

In addition to static feature importances, we looked into if the forecasting errors of SSM-CGM differ across diabetes status, age, and clinical site (Figure \ref{fig:StaticErrors}). We find that on average, people with diabetes (either insulin or oral medications) have higher mean absolute error (MAE) than people who are healthy or have prediabetes. Additionally, we suspect people with diabetes taking insulin have a larger MAE in the SSM-CGM model since the more unpredictable glucose dynamics are attributed to not having information of when individuals take insulin in the AI-READI dataset. On average, we see people who are older tend to have a higher MAE but the effect is small (r = 0.09, p-val = 0.019). For clinical site, we observe that the global average MAE is within the confidence bands of the sites used participant recruitment. This is promising as clinical site is an observable batch effect feature in the AI-READI dataset.

\newpage
\subsection{Interpretability - Embeddings}\label{app:embeddings}

\begin{figure}[h]
    \centering 
    \includegraphics[width=0.85\textwidth]{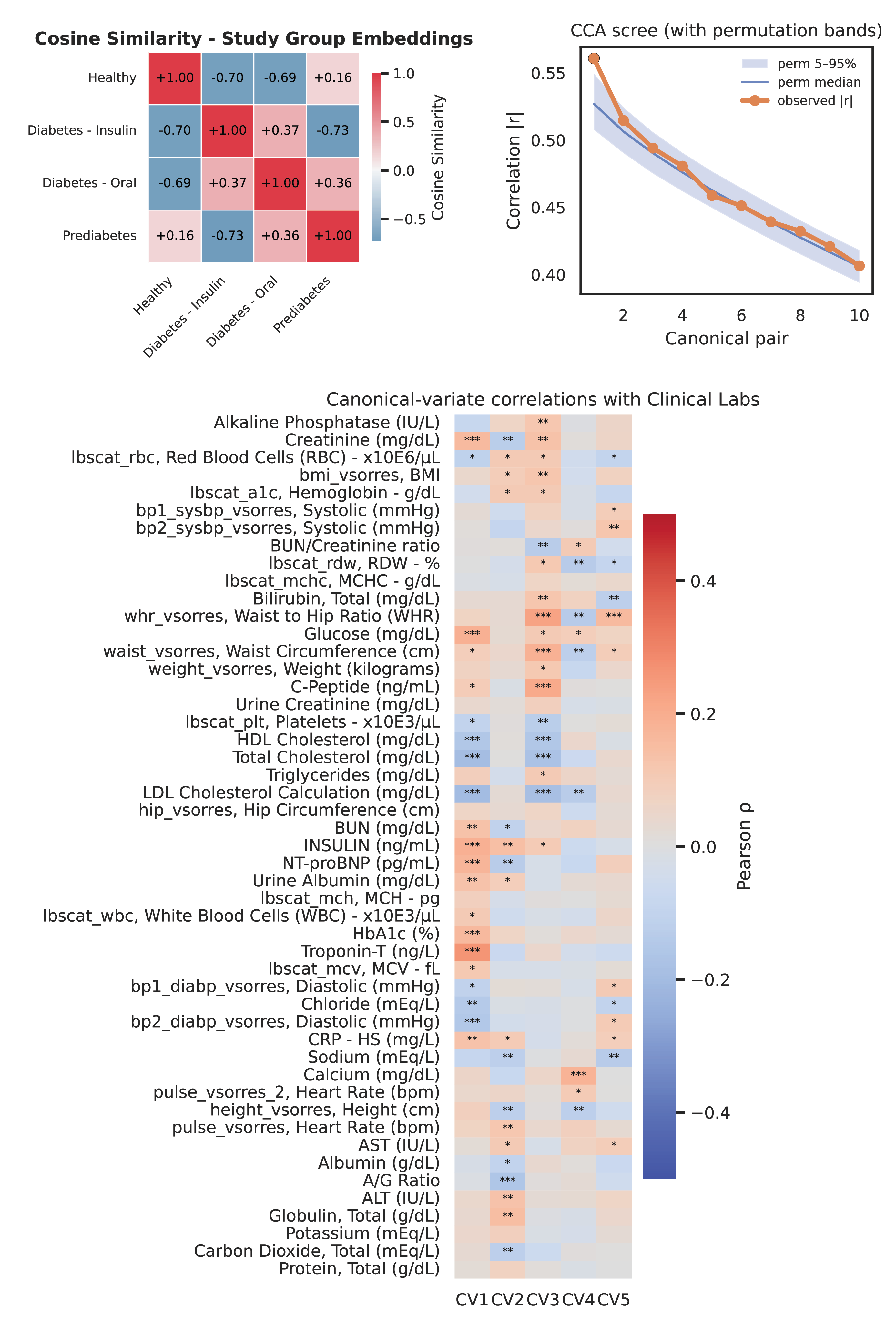}
    \caption{Interpreting participant ID embeddings of SSM-CGM.}
    \label{fig:Embeddings}
\end{figure}

To reason whether SSM-CGM captures glucose dynamics we utilized the static embeddings of diabetes status and participant ID (Figure \ref{fig:Embeddings}). Cosine similarity revealed that embeddings of healthy individuals were distinct from those of individuals with diabetes, while people with diabetes on different medications (insulin vs. oral) shared similar embeddings (sim = 0.37). Individuals with prediabetes were closer to healthy participants than to those with diabetes on either oral or insulin therapy. This ordering reflects clinical knowledge: healthy and individuals with prediabetes generally maintain better glucose control than people with type 2 diabetes, with those on insulin, typically a last-line treatment, having the most difficult-to-control glucose \cite{DiabMedADA2018}.

To further understand what SSM-CGM is learning, we took the individual static embeddings and utilized canonical correlation analysis (CCA) to compare if the variation in individual embeddings share similar axes of variation with standard clinical lab values (Figure \ref{fig:Embeddings}). For robustness, we utilized permutation testing (n = 10,000) samples to identify a null distribution of expected range of correlations under no true association. We observed that the 1st canonical pair has an observed canonical correlation above the null confidence interval suggesting a robust signal between SSM-CGM individual embeddings with clinical lab features, while the 2nd-5th canonical pairs display correlations that might be observed by random chance. Based on this finding, we correlated clinical lab values with the canonical-variates of the SSM-CGM embedding to observe which lab values are corresponding to each canonical pair CV1-5. For CV1, troponin, HbA1c, insulin, glucose, and cholesterol features tend to be explain the signals shared between the individual embeddings captured and the clinical lab values. This variation likely reflects glycemic traits and characteristic metabolic signatures associated with diabetes \cite{Ahlqvist2018-sa}.

\newpage
\subsection{Identification Assumptions for Counterfactual Forecasting}\label{app:plausibility}

\paragraph{Objective.} Counterfactual forecasting in our work corresponds to planned interventions on future covariates (e.g., heart rate, respiration). Formally, the goal is to estimate how future glucose trajectories would change if we were to set these covariates to specified values, irrespective of their natural dynamics.

\paragraph{Sequential g-formula.} 
Under assumptions of consistency, sequential ignorability, and positivity, 
the counterfactual distribution of a planned covariate sequence 
$\mathbf{a}_{t+1:t+h}$ is identified by the sequential \emph{g-formula}:
\begin{equation}
p(\mathrm{CGM}_{t+1:t+h}(\mathbf{a}) \mid \mathcal{H}_t)
= \prod_{s=t+1}^{t+h} p(\mathrm{CGM}_s \mid \mathcal{H}_{s-1}, X_s = a_s),
\end{equation}
where $\mathcal{H}_{s-1}$ is updated with simulated 
$\mathrm{CGM}_{t+1:s-1}$ and fixed $X_{t+1:s-1}=\mathbf{a}_{t+1:s-1}$ 
under this recursive definition.

In our work, \textbf{SSM-CGM} directly conditions on the entire planned sequence 
$\mathbf{a}_{t+1:t+h}$ in a single forward pass, yielding
\begin{equation}
\widehat{p}_\theta(\mathrm{CGM}_{t+1:t+h} \mid \mathcal{H}_t, \mathbf{a}_{t+1:t+h}),
\end{equation}
which is \emph{equivalent in principle} to the g-formula factorization but avoids 
explicit recursive simulation.

\paragraph{Identification Assumptions.}
Identification of counterfactual CGM forecasts requires the standard assumptions 
of \emph{consistency}, \emph{sequential ignorability}, and \emph{positivity}. While these assumptions are formally untestable from observational data, we provide supporting evidence and domain-specific reasoning suggesting 
they are plausible in our setting. 
For a deterministic future intervention plan $\mathbf{a}_{t+1:t+h}$, these can be written 
in sequence-level notation as follows:

\paragraph{Consistency.} 
If the observed covariates equal the planned intervention sequence, then the observed outcome 
trajectory equals the counterfactual trajectory:
\begin{equation}
\text{If } X_{t+1:t+h} = \mathbf{a}_{t+1:t+h}, 
\quad \text{then } \mathrm{CGM}_{t+1:t+h}(\mathbf{a}) = \mathrm{CGM}_{t+1:t+h}.
\end{equation}

Consistency requires that when the planned intervention path coincides with the observed covariates, the counterfactual glucose trajectory equals the observed trajectory. Consistency is supported by the use of objective sensor-based measurements (e.g. heart rate, respiration rate, glucose levels, etc.), which ensure that observed and planned values are defined on the same measurement scale. Our counterfactual forecasting procedure implements this assumption directly. When the planned action sequence equals the observed one, the counterfactual forecast reduces to the model’s standard prediction of the observed trajectory. Although forecasts do not need to equal the realized trajectory exactly, this construction aligns the model with the formal consistency assumption.

\paragraph{Sequential Ignorability (No Unmeasured Confounding).} 
Conditional on the observed history, future counterfactual glucose trajectories are independent of planned actions:
\begin{equation}
\mathrm{CGM}_{t+1:t+h}(\mathbf{a}) \;\perp\!\!\!\perp\; 
\mathbf{a}_{t+1:t+h} \;\mid\; \mathcal{H}_t.
\end{equation}

This assumption requires that, conditional on the observed history $\mathcal{H}_t$, the planned intervention $X_s$ is independent of future counterfactual glucose trajectories. 
In our setting, this assumption is more plausible because we treat HR, RR, and activity measures as the direct intervention variables. By intervening directly on these sensor-based covariates, rather than on upstream unobserved causes (e.g., fitness, hormones), we reduce the scope for hidden confounding. The inclusion of rich histories (CGM, activity, meals, HR) further mitigates omitted-variable bias, making sequential ignorability more credible in this framework.

\begin{figure}[h]
    \centering 
    \includegraphics[width=0.9\textwidth]{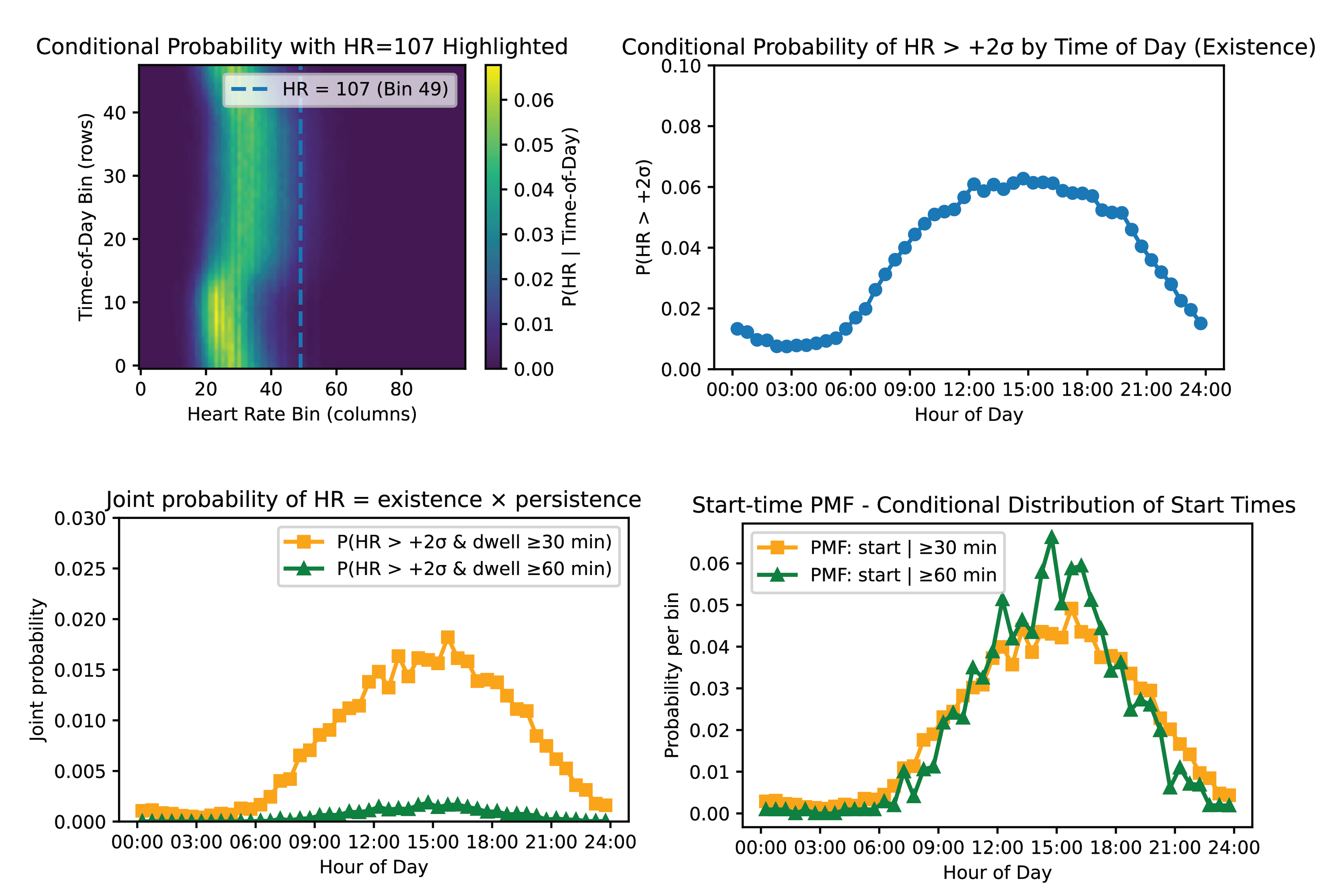}
    \caption{Positivity of elevated heart rate (HR) across the day.}
    \label{fig:HRplausible}
\end{figure}

\begin{figure}[h]
    \centering 
    \includegraphics[width=0.9\textwidth]{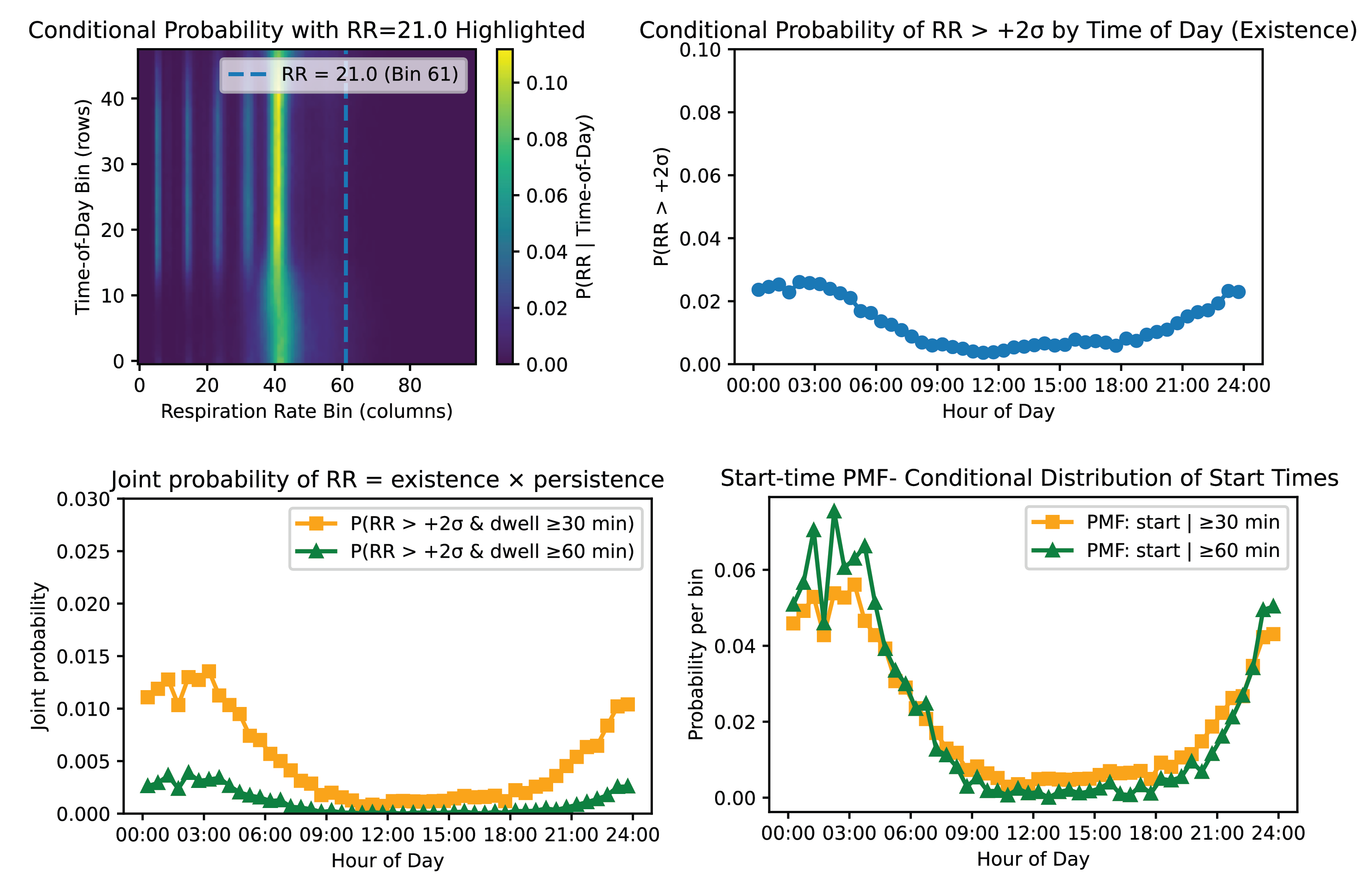}
    \caption{Positivity of elevated respiration rate (RR) across the day.}
    \label{fig:RRplausible}
\end{figure}

\paragraph{Positivity.} 
For any history that occurs with positive probability, all planned actions must have positive 
probability of being observed:
\begin{equation}
\Pr(\mathbf{a}_{t+1:t+h} \mid \mathcal{H}_t) > 0 
\quad \text{whenever } p(\mathcal{H}_t) > 0.
\end{equation}

We empirically assess positivity, by examining the distributions of two representative covariates, heart rate (HR; Figure \ref{fig:HRplausible}) and respiration rate (RR; Figure \ref{fig:RRplausible}), across times of day in the AI-READI cohort. Both covariates showed support across physiologically plausible ranges, with conditional probabilities $p(a_{t+1:t+h} \mid \text{Time of Day}) > 0$ throughout. For example, a $+2$ SD HR perturbation (107 bpm) was most commonly observed between 9~AM and 9~PM, consistent with daytime activity, while a $+2$ SD RR perturbation (21 breaths/min) was most supported between 9~PM and 6~AM, reflecting nighttime respiration. These patterns demonstrate that our counterfactual interventions, restricted to times of day where such values are commonly observed, lie within empirically supported regions of covariate space. Importantly, these distributions also show persistence: elevated HR or RR lasting 30–60 minutes remains plausible (Figures \ref{fig:HRplausible}, \ref{fig:RRplausible}). Therefore, moderate shifts are well-represented across contexts, satisfying positivity, whereas extreme or physiologically implausible regimes (e.g., very high HR during sleep) would violate positivity and are excluded from analysis.

\subsection{Limitations and Future work}\label{app:limitations}
Our study has several limitations spanning the dataset, modeling framework, interpretability, and clinical scope.

\paragraph{Dataset-related limitations.} The AI-READI dataset lacks direct annotations for meals and medication use (e.g., insulin, metformin, GLP1), both of which strongly influence glucose dynamics. We addressed the absence of meal records by training a meal-detection model on an external dataset, but this provides only a coarse proxy for meal timing and does not capture meal size or composition. Improving meal detection through higher-quality annotations will be important for refining future models. Similarly, unrecorded insulin administration introduces variability that likely contributes to higher prediction error in insulin-treated participants.

\paragraph{Modeling and benchmarking limitations.} Although SSM-CGM consistently outperformed the Temporal Fusion Transformer (TFT), our benchmarking relied on TFT as the main comparator. Additional comparisons to alternative forecasters and broader ablation studies, including the roles of future covariates, preprocessing, and alternative Mamba configurations (e.g., ablation of light vs. stacked layers) will clarify which model components drive performance gains. We also note that our runtime is substantial (1.5 hour per epoch on 4 A100 GPUs), and further work is needed to improve efficiency.

\paragraph{Interpretability and counterfactual reasoning.} While the model provides interpretable outputs via variable selection and temporal attribution, these remain associational rather than causal. Counterfactual forecasts are identified under standard assumptions of consistency, sequential ignorability, and positivity, which cannot be fully validated from observational data. As such, interpretability and counterfactual outputs should be viewed as hypothesis-generating rather than definitive.

\paragraph{Forecasting scope.} Our evaluation was limited to a 1-hour forecasting horizon. While this window aligns with near-term clinical decision-making, extending to forecasts of different lengths could support a wider range of use cases. Moreover, our current counterfactual interventions targeted single covariates, whereas real-world behaviors involve coordinated, multivariate activity patterns. Extending counterfactual reasoning to multivariate interventions could better support actionable, individualized recommendations.

\paragraph{Generalization across cohorts.} Our results are specific to the AI-READI cohort, and it remains unclear how well the model generalizes across datasets with different demographics, devices, or recording conditions. The effective sample size for detecting model improvements is not well established, and replication in other large-scale CGM cohorts will be essential. 

\paragraph{Future Directions.} Beyond addressing these limitations, several extensions are possible. Incorporating planned categorical covariates (e.g., scheduled exercise, meals, or sleep) could make forecasting and counterfactuals more realistic, since individuals can plan activities but not continuous physiological values such as heart rate minute by minute. Extending the same model architecture to multivariate forecasting of both glucose and wearable signals could strengthen counterfactual reasoning by providing forecasted future inputs on learned behavior. Building streaming-capable versions of SSM-CGM, leveraging the linear-time properties of Mamba, would improve deployability. Finally, leveraging SSM-CGM forecasting models combined with reinforcement learning methods could recommend personalized, adaptive behavioral policies to maximize time-in-range (70–180 mg/dL).

\subsection{Code Availability}



Code is available at \href{https://github.com/shakson-isaac/SSM-CGM}{github.com/shakson-isaac/SSM-CGM} under the Harvard license for non-commercial research use.

\end{document}